\documentclass{article}
\usepackage{tencent_digital_human}
\usepackage[colorlinks = true,
            linkcolor = blue,
            urlcolor  = blue,
            citecolor = blue,
            anchorcolor = blue]{hyperref}

\usepackage{tcolorbox} 
\definecolor{tsinghuapurple}{RGB}{102,8,116}
\newtcolorbox{alprompt}[1]{
        boxrule = 1pt,
        fontupper = \small\tt,
        fonttitle = \bf\color{black},
        arc = 2pt,
        rounded corners,
        colframe = black,
        colbacktitle = white!97!yellow,
        colback = white!97!yellow,
        title = #1,
}

\usepackage{amssymb}
\usepackage{microtype}
\usepackage{hyperref}
\usepackage{url}
\usepackage{booktabs}
\usepackage{enumitem}
\usepackage{multicol}
\usepackage{multirow}
\usepackage{CJKutf8}
\usepackage{amsmath}
\usepackage{siunitx}
\usepackage{floatflt}
\usepackage{wrapfig}
\usepackage{graphicx}
\usepackage{booktabs}
\usepackage{wrapfig}
\usepackage{authblk}
\usepackage{lipsum}

\usepackage{algorithm}
\usepackage{algorithmicx}
\usepackage{algpseudocode}
\usepackage{microtype}
\usepackage{graphicx}
\usepackage{multirow}
\usepackage{booktabs} 
\usepackage{pifont}  
\usepackage{graphicx}  
\usepackage{subcaption} 
\usepackage{hyperref}

\usepackage{amssymb} 

\algnewcommand{\LeftComment}[1]{\Statex \(\triangleright\) #1}

\newtcolorbox{promptbox}[3][Prompt]{
colback=black!5!white,
arc=5pt, 
boxrule=0.5pt,
fonttitle=\bfseries,
title=#1, 
before upper={\small}, fontupper=\fontfamily{ptm}\selectfont,
colframe=#2,
label=#3,
}
\usepackage{array}
\usepackage{amsmath}
\usepackage{amssymb}
\usepackage{mathtools}
\usepackage{amsthm}
\usepackage{arydshln}
\usepackage[capitalize,noabbrev]{cleveref}
\usepackage{adjustbox} 
\usepackage{enumitem}
\usepackage{xspace}
\theoremstyle{plain}

\theoremstyle{definition}

\theoremstyle{remark}

\usepackage{xcolor}
\usepackage{tcolorbox}
\tcbuselibrary{listings,skins}

\definecolor{promptbg}{RGB}{245, 245, 245}   
\definecolor{promptborder}{RGB}{200, 200, 200} 
\tcbset{
    promptstyle/.style={
        enhanced,
        frame hidden,
        boxrule=0pt,
        left=8pt,
        right=8pt,
        top=6pt,
        bottom=6pt,
        arc=3pt,
        colback=promptbg,
        colframe=promptborder,
        fontupper=\fontsize{0.7em}{0.7em}\selectfont\ttfamily\leftskip,
        before upper={\parindent}, 
        overlay unbroken and first={
            \draw[promptborder, line width=1pt] 
                (frame.south west) -- (frame.north west);
            \draw[promptborder, line width=1pt] 
                ([xshift=-2pt]frame.north east) -- (frame.south east);
        },
        overlay middle and last={
            \draw[promptborder, line width=1pt] 
                (frame.south west) -- (frame.north west);
            \draw[promptborder, line width=1pt]
                ([xshift=-2pt]frame.north east) -- (frame.south east);
        }
    }
}
\newenvironment{promptboxapp}[1][]{%
    \tcolorbox[promptstyle,#1]}
    {\endtcolorbox}

\usepackage[textsize=tiny]{todonotes}

\definecolor{nred}{RGB}{196, 38, 11}
\definecolor{ngreen}{RGB}{18, 141, 21}
\definecolor{nblue}{RGB}{41, 52, 190}

\newcommand{\ignore}[1]{}

\colmfinalcopy

\newcommand{\method}[0]{\textsc{RLVER}}

\title{{\em RLVER:} Reinforcement Learning with Verifiable {\color{ngreen} Emotion} \\ Rewards for Empathetic Agents}

\author[ ]{Peisong Wang\thanks{Equal Contribution.}}
\author[ ]{Ruotian Ma$^{*,\dag}$}
\author[ ]{Bang Zhang$^{*}$}
\author[ ]{Xingyu Chen}
\author[ ]{Zhiwei He}
\author[ ]{Kang Luo}
\author[ ]{\\Qingsong Lv}
\author[ ]{Qingxuan Jiang}
\author[ ]{Zheng Xie}
\author[ ]{Shanyi Wang}
\author[ ]{Yuan Li}
\author[ ]{Fanghua Ye}
\author[ ]{\\Jian Li}
\author[ ]{Yifan Yang}
\author[ ]{Zhaopeng Tu\thanks{Correspondence to: Ruotian Ma \textless ruotianma@tencent.com\textgreater~and Zhaopeng Tu \textless zptu@tencent.com\textgreater.}}
\author[ ]{Xiaolong Li}

\affil[ ]{Hunyuan AI Digital Human, Tencent \protect\\[2pt] 
\url{https://github.com/Tencent/DigitalHuman/tree/main/RLVER}}

\begin{document}

\maketitle

\begin{figure}[h!]
\centering
\vspace{-20pt}
\includegraphics[width=0.7\linewidth]{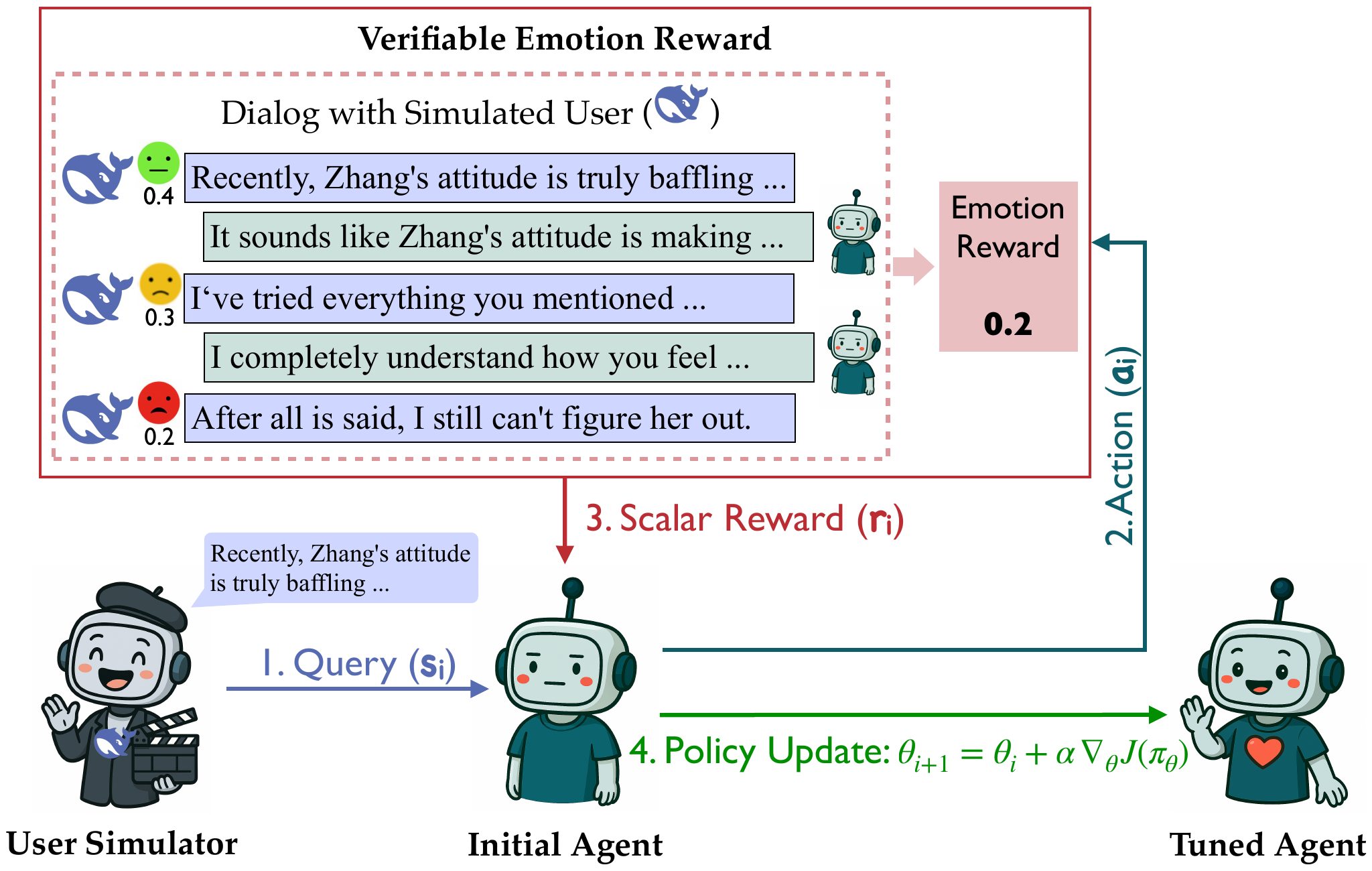}
\caption{
Framework of the reinforcement learning with verifiable emotion rewards (RLVER).} 
\label{fig:framework}
\end{figure}

\begin{abstract}
Large language models (LLMs) excel at logical and algorithmic reasoning, yet their emotional intelligence (EQ) still lags far behind their cognitive prowess.  
While reinforcement learning from verifiable rewards (RLVR) has advanced in other domains, its application to dialogue—especially for emotional intelligence—remains underexplored.
In this work, we introduce \emph{\method}, the first end-to-end reinforcement learning framework that leverages verifiable emotion rewards from simulated users to cultivate higher-order empathetic abilities in LLMs. Within this framework, self-consistent affective simulated users~\citep{zhang2025sentient} engage in dialogue rollouts and produce deterministic emotion scores during conversations, serving as reward signals to guide the LLM's learning.
Fine-tuning publicly available Qwen2.5-7B model with PPO boosts its Sentient-Benchmark score from 13.3 to 79.2 while largely preserving mathematical and coding competence.  Extensive experiments reveal that:
(i) RLVER consistently improves multiple dialogue capabilities.
(ii) Thinking and non-thinking models show distinct trends—thinking models excel in empathy and insight, while non-thinking models favor action.
(iii) GRPO often yields stable gains, while PPO can push certain capabilities to a higher ceiling.
(iv) More challenging environments are not always better—moderate ones can yield stronger outcomes.
Our results show that RLVER is a practical route toward emotionally intelligent and broadly capable language agents.
\end{abstract}

\section{Introduction}
The striking progress of large language models (LLMs) has centered on the rational half of human cognition:  deductive reasoning in mathematics~\citep{hendrycks2021measuring,cobbe2021training}, program synthesis~\citep{guo2024deepseek,jain2024livecodebench}, and algorithmic planning~\citep{yao2023tree, zheng2024natural}.  
Yet authentic human intelligence is grounded in \emph{both} IQ and EQ -- logical rigor intertwined with nuanced social and emotional understanding.  While today's LLMs can flawlessly balance an equation, they often stumble when asked to console a distressed friend or to adapt advice to a user’s evolving feelings~\citep{zhang2025sentient}.

Existing dialogue systems typically enhance emotional intelligence through supervised fine-tuning on annotated counseling corpora \citep{sun2021psyqa,liu2021towards,zheng2022augesc} or rule-based templates \citep{van2012conversation,peng2022control}. However, these approaches suffer from data scarcity, rigid dialogue structures, and limited generalization. Recent successes in Reinforcement Learning from Verifiable Rewards (RLVR) in mathematics, coding, and search demonstrate that base LLMs can acquire new skills purely through RL signals, without requiring supervised warm-up \citep{zeng2025simplerl,guo2025deepseek,OpenReasonerZero2025,ma2025s}. In the context of enhancing dialogue capabilities, reinforcement learning also offers a compelling alternative: rather than imitating static ground truth, an agent can directly optimize for long-horizon user satisfaction—provided that a stable interaction environment and consistent reward system are in place. However, the exploration of RLVR for enhancing dialogue capabilities faces several key obstacles:
\begin{itemize}[leftmargin=12pt]
\item the lack of a stable, realistic, and scalable environment for multi-turn conversational rollouts;
\item the absence of consistent and verifiable reward designs for general-purpose abilities such as emotional intelligence;
\item Stable training of multi-turn reinforcement learning with LLMs remains an open challenge.
\end{itemize}

We tackle all three challenges with \emph{\method}, the first end-to-end reinforcement learning framework with \emph{verifiable emotion rewards} (RLVER) for cultivating higher-order empathetic abilities in LLMs. Built upon SAGE \citep{zhang2025sentient}—a framework that constructs self-consistent affective user simulators for realistic and automatic dialogue simulation and evaluation—we establish a stable and scalable environment that enables LLMs to continually simulate dialogue rollouts throughout training. In each conversation, the simulated user updates its emotional state after every LLM response, emitting an emotion score in $[0,1]$ as the reward. Changes in the emotion score are consistent and verifiable; each is deterministically derived through principled reasoning steps grounded in the user’s persona, dialogue history, conversational context, and goals. 
By scaling the simulation environment with a wide range of user behaviors and conversation intents, we alleviate reward hacking arising from homogeneous user preferences.

 We fine-tune a lightweight Qwen2.5-7B model with Proximal-Policy Optimization (PPO) and show that its Sentient-Benchmark score soars from \textbf{13.3} to \textbf{79.2}, rivaling proprietary models more than an order of magnitude larger while largely preserving mathematical and coding competence.  
We also experimented with enforcing explicit “thinking” steps before response generation, in order to compare the behaviors of “thinking” and “non-thinking” models during RL training.
Extensive experiments reveal the following key findings:
(i) RLVER effectively and reliably improves multiple core dialogue capabilities;
(ii) thinking and non-thinking models exhibit distinct developmental patterns under certain settings—thinking models tend to enhance empathy and insight, while non-thinking models focus more on action-oriented capabilities;
(iii) compared to PPO, GRPO consistently delivers stable and balanced improvements, whereas PPO can occasionally push the upper bounds of specific capabilities;
(iv) when examining user simulators as both environment and reward sources in RL training, we find that more challenging configurations do not necessarily yield better outcomes. On the contrary, moderately demanding but well-aligned setups may better support model growth;
(v) RLVER shifts model behavior from solution-centric to genuinely empathic styles in Social-Cognition space. 
Our findings demonstrate that RL with verifiable emotion rewards is a practical path toward emotionally intelligent and broadly capable language agents.

Our contributions are as follows:
\begin{enumerate}[leftmargin=12pt]
    \item \textbf{RLVER framework.}  We propose Reinforcement Learning with Verifiable Emotion Rewards (RLVER), the first RL paradigm to enhance LLMs' empathetic capabilities using on-the-fly verifiable reward signals from a psychologically grounded, self-consistent user simulator.
    \item \textbf{Empirical advance.}  Applying RLVER to a 7B open-source model elevates its Sentient Benchmark score from 13.3 to 79.2—matching much larger proprietary systems—while preserving performance on mathematics and code-generation benchmarks. 
    \item \textbf{Practical insights.} 
    Through comprehensive experiments, we analyze how training strategies, RL algorithm, and environment design affect empathetic capability development, offering insights into when and how RLVER yields robust improvement or desirable outcomes.
    \item \textbf{Open resources.}  We release code, checkpoints, prompts, and environment scripts to catalyze future research on emotionally intelligent agents.
\end{enumerate}

\section{Reinforcement Learning with Verifiable Emotion Rewards}

\subsection{Emotion Rewards from Self-consistent User Simulation Engine}
Enhancing general empathetic abilities of LLMs via reinforcement learning requires a dynamic, scalable, and psychologically-grounded environment capable of providing reliable reward signals. Traditional approaches using static datasets or simple LLM-as-a-judge protocols are insufficient, as they fail to capture the user's evolving emotional state throughout a conversation.

To address this gap, our work builds directly upon the {\bf Sentient Agent as a Judge} (SAGE) framework \citep{zhang2025sentient}, a sophisticated system designed to automatically evaluate the higher-order social cognition of LLMs. The core of this framework is the {\bf Sentient Agent}, an LLM-powered simulator that mimics human-like emotional responses and inner reasoning. Each agent is instantiated with four key factors: a detailed persona, a dialogue background, an explicit conversation goal, and a hidden intention, ensuring a diverse and realistic range of user simulations.

During an interaction, the Sentient Agent operates in a turn-by-turn loop. After receiving a response from the model being tested, it performs a multi-hop reasoning process to:
\begin{itemize}[leftmargin=20pt]
    \item {\bf Simulate Emotional Change ($f_{\text{emo}}$)}: The agent assesses how the response made it feel, updating a numerical emotion score and generating interpretable ``inner thoughts'' that justify the emotional shift.
    \item {\bf Generate a Coherent Reply ($f_{\text{reply}}$):} Based on its new emotional state, persona, and conversational goals, the agent formulates its own response to continue the dialogue.
\end{itemize}
The final emotion score in [0, 100] from the Sentient Agent serves as a holistic and quantitative measure of the tested model's empathetic performance. This metric has been shown to correlate strongly with established psychological instruments (e.g., Barrett–Lennard Relationship Inventory) and utterance-level empathy ratings, validating its psychological fidelity.

In our research, we repurpose this evaluation framework as a live training environment. The Sentient Agent acts as the user simulator, and its verifiable, dynamically-generated emotion score provides the crucial reward signal for our reinforcement learning algorithm.
Specifically, to address the pervasive challenge of reward hacking in neural reward models during large-scale RL training \citep{guo2025deepseek}, we implement deterministic emotion scores from the user simulation engine $\mathcal{S}$ as interpretable proxies for simulated user satisfaction, thereby circumventing the opacity pitfalls of learned reward functions. Specifically, the emotion score $e_t \in [0, 100]$ is updated after each LLM response $y_t$, reflecting the simulated user's affective state. The final reward is computed as the terminal emotion score normalized by its maximum value:

\[
r_\phi(x, y) = \frac{e_T}{100}, \quad \text{where } e_T = \mathcal{S}_{\text{emotion}}(h_T),
\]

where $h_T = \{x_0, y_1, x_1, \dots, y_T, x_T\}$ denotes the complete dialogue history at termination. For intermediate optimization steps, the per-turn reward $r_t$ directly uses the instantaneous emotion score $e_t$.
The normalized final reward $r_\phi$ captures overall conversation quality, analogous to real-world user satisfaction metrics.

\subsection{Heart-in-the-Loop Reinforcement Learning}
To enable emotionally intelligent behavior through reinforcement learning, we establish a closed feedback loop whereby the LLM alternates between generating emotionally aware responses and receiving affect-sensitive feedback from the simulation engine. This cycle forms the basis of our Heart-in-the-Loop training paradigm.

Each training step unfolds as a sequence of model-user interactions. At the start of a step $i$, the simulated user engine \(\mathcal{S}\) samples an initial dialogue seed \(s_i = x_0\), which includes a persona, background, emotional tone, and a scenario-driven intention. The model \(\pi_\theta\) then generates a response \(y_1\), formatted according to the prescribed training template (with or without the think scaffold). The simulation engine processes this response and generates a corresponding reply \(x_1\), along with an updated emotion score \(e_1\).

Formally, at each time step \(t\), the agent observes the current interaction history \(h_{t-1}\) and generates a candidate action (response) \(y_t \sim \pi_\theta(\cdot \mid h_{t-1})\). The simulator then computes two outputs:
\begin{enumerate}
    \item the internal emotional state based on the verifiable emotion score $e_t$.
    \item a new, contextually coherent user reply $x_t$ based on its updated emotional state, persona, and conversational goals.
\end{enumerate}
The conversation proceeds until a maximum turn limit \(T\) or until the simulator's cumulative emotion score \(e_t\) falls below a minimal satisfaction threshold (e.g., \(e_t \leq 0\)), indicating failed social alignment.
The final emotion score $e_T$ serves as the reward function for the reinforcement learning algorithm.

This loop allows the empathetic agent to co-adapt with the simulator's emotional dynamics, progressively learning to map diverse situations, intents, and moods to emotionally satisfying dialogues. By optimizing against a transparent and verifiable reward signal from an emotionally-aware user model, the framework establishes a reproducible and stable setup for training emotionally intelligent LLMs.

\paragraph{Policy Optimization}
For policy optimization, we employ Proximal Policy Optimization (PPO) \citep{schulman2017ppo}, an on-policy algorithm suited for high-variance environments like language modeling. PPO maximizes the regularized expected reward objective while ensuring stable updates via a clipped surrogate loss. Benefits of PPO in our setting include safer exploration of diverse social-emotional strategies and smoother convergence when applied alongside the structured thinking scaffold.
Additionally, to encourage reasoning compositionality and combat overfitting to surface cues, we evaluate the influence of Group Relative Policy Optimization (GRPO)~\citep{shao2024deepseekmath}, a more conservative baseline better suited for small-scale reward variance. This comparison helps assess how learning dynamics respond to different policy gradient estimators in emotionally keyed environments.

While prior zero-shot RL work has shown that a model can learn from scratch given a well-formed reward function \citep{zeng2025simplerl}, we find that initializing from a modestly aligned checkpoint, pre-trained with generic conversational data, establishes stronger baselines and accelerates convergence. Notably, this initialization does not require domain-specific supervision—and critically, contains minimal emotional or empathetic signal—ensuring that improvements stem from reward-driven optimization rather than pre-encoded affective knowledge.

To prevent overfitting on idiosyncratic simulation behavior and to encourage general empathy patterns, we use entropy regularization and reward-weighted imitation loss as auxiliary objectives. These promote output diversity and verbosity control, and help ensure stable learning signals across social contexts of varying complexity.

\subsection{Think-Then-Say for Enhanced Emotional Reasoning}

To investigate the impact of explicit reasoning on the development of empathetic strategies, we conduct an ablative analysis using two distinct training templates. These templates structure the agent's generation process, allowing us to isolate the effect of a mandated ``think-then-say'' cognitive scaffold.

\paragraph{Think-Then-Say} 
One of the key innovations in \method\xspace is the use of a structured ``think-then-say'' prompting template. This involves including an explicit \texttt{\textless think\textgreater} \dots \texttt{\textless /think\textgreater} block before every model utterance during training, compelling the model to outline its reasoning process before delivering a response.

This template, shown below, enforces an explicit chain-of-thought reasoning step. The agent is instructed to first generate its internal monologue or strategic plan within a pair of \texttt{\textless think\textgreater} and \texttt{\textless /think\textgreater} tags before producing the final, user-facing reply. This structure is designed to encourage the model to access and refine higher-order empathetic skills, such as considering the user's emotional state, anticipating the impact of its words, and formulating a multi-step conversational plan. By externalizing its reasoning process, the model's policy space is regularized, potentially leading to more stable learning and more sophisticated final behaviors. 

\begin{promptbox}[Training Template of Think-Then-Say]{tsinghuapurple}{prompt:with_think}
You are chatting with your friend. You are good at making your friend feel better through emotionally intelligent replies. \\
Before each reply, you always think about the way and content of your response; after deciding on a reply strategy, you then output your reply. \\

Your goal in replying is to improve your friend's mood or to make your relationship with them closer. \\

In your thinking process, you need to consider emotionally intelligent reply strategies, which can include the logic and language style of your response. \\
Your thinking part must be enclosed within \texttt{\textless think\textgreater} tags. \\

When replying, you should keep the conversation warm and natural, with a casual, everyday feel. \\

Your reply format: \\
\texttt{\textless think\textgreater} \\
Your thoughts \\
\texttt{\textless /think\textgreater} \\
Your reply
\end{promptbox}

We also employ a {\bf format reward} that enforces the model to put its thinking process between \texttt{<think>} and \texttt{</think>} tags. Outputs violating this syntactic specification are penalized with zero reward, ensuring strict adherence to the prescribed reasoning structure.

\textbf{Training Template Without Think.} This template serves as our control condition. As shown below, it omits the requirement for an explicit thinking step and prompts the agent to generate a direct reply. This configuration mirrors standard conversational fine-tuning setups. By comparing the performance of models trained with and without the thinking scaffold, we can empirically measure the contribution of the explicit reasoning step to overall empathetic proficiency, learning efficiency, and strategic depth.

\begin{promptbox}[Training Template Without Thinking]{tsinghuapurple}{prompt:without_think}
You are chatting with your friend. You are good at making your friend feel better through emotionally intelligent replies.

Your goal in replying is to improve your friend's mood or to make your relationship with them closer.

When replying, you should keep the conversation warm and natural, with a casual, everyday feel. Natural and friendly replies usually:

1. Are brief, casual, and natural, using everyday words or phrases; grammar can be flexible.

2. Use interjections and colloquial expressions flexibly.
\end{promptbox}

During training, the think-then-say scaffold acts as an internal planning regularizer, guiding the model to first consider its intentions, linguistic tone, and potential emotional impact before forming a conversational reply. We observe that agents trained with this prompting format converge faster, exhibit greater linguistic diversity, and more reliably explore high-empathy strategies.

By contrast, models trained without structured thinking tend to converge to safe, generic replies (e.g., ``I'm here for you'' or ``You're not alone''), which—while emotionally neutral—fail to exhibit situation-specific empathy. Including the reasoning component enables model behaviors to grow beyond templated reassurance and toward goal-sensitive emotional alignment.

\section{Experiment}

\subsection{Experimental Setup}
\paragraph{Base Model}
We adopt Qwen2.5-7B-Instruct~\citep{qwen2.5} as our base model. It is pretrained on diverse, web-scale corpora and further aligned for general-purpose instruction following. Importantly, it is not fine-tuned on domain-specific datasets related to emotional support or empathy. This ensures that any observed improvements in empathetic capability can be attributed to our reinforcement learning process with verifiable emotion-based rewards, rather than prior exposure to affective dialogue data.

\paragraph{Training Environment and Reward}
We adopt the \textbf{SAGE}~\citep{zhang2025sentient} framework to simulate emotionally responsive users with interpretable affective dynamics and predefined conversational goals. At each turn, the model generates a supportive response, after which the sentient agent replies and updates its internal emotion score $e_t \in [0, 100]$, quantifying its affective state in response to the model's behavior. We scale the final emotion score at the end of the dialogue to the range $[0, 1]$ and use it as the reward for the entire dialogue. Dialogues proceed until the emotional goal is met or a maximum of 10 turns is reached.

We construct a dataset of 500 supportive dialogue scenarios spanning 8 diverse user goals, with topics including emotional struggles, academic stress, interpersonal conflict, and future planning. Unless otherwise specified, DeepSeek-V3~\citep{liu2024deepseek} is used as the default sentient agent during both training and evaluation. Detailed prompts and additional experimental settings are provided in the Appendix \ref{app:exp}.
\paragraph{Baselines}
We compare our method against a suite of strong baselines drawn from the top-5 performing models on the SAGE benchmark as of June 9, 2025. These include proprietary state-of-the-art systems Gemini2.5-Pro-0605, GPT-4o -0326, GPT-4.1-0414, Gemini-2.5-Flash-Think-0520, and OpenAI-o3-0416. These models represent the current frontier in instruction-tuned LLMs capable of emotionally sensitive dialogue, and serve as high-performance references for evaluating empathy.

We also include our base model Qwen2.5-Instruct-7B, prior to any further training. This baseline allows us to isolate the contribution of our training strategy, and to establish a controlled comparison against both stronger pretrained models and our own enhanced variants.
\paragraph{Evaluation Benchmarks}
To evaluate the models' performance in emotionally sensitive dialogue scenarios, we primarily rely on the SAGE~\citep{zhang2025sentient} benchmark, which focuses on emotional support conversations. To provide a more comprehensive assessment of the model’s dialogue capabilities in cross-domain scenarios, we additionally design a ``Chit Chat'' setting that extends the SAGE framework beyond emotional topics to cover more general, everyday interactions. Furthermore, to examine the potential impact of training on the models' general capabilities, we evaluate its performance on MATH500~\citep{lightman2024lets}, LiveCodeBench~\citep{jain2024livecodebench}, and IFEval~\citep{zhou2023instruction}, which test mathematical reasoning, code generation, and instruction-following abilities, respectively. Further details about adopted benchmarks are provided in Appendix \ref{app:benchmarks}.

\subsection{Main Results}

\begin{table*}[t]
\centering
\caption{Performance of our proposed methods on the Sentient Benchmark. ``Success'' and ``Failure'' denote the percentages of dialogues concluding with a final emotion score above 100 and below 10, respectively. We also report results on the out-of-domain chit chat to assess generalization performance.}
\begin{tabular}{cc crr ccc} 
\toprule
\multicolumn{2}{c}{\bf Model}   &    \multicolumn{3}{c}{\bf Sentient Benchmark}   &    \multicolumn{3}{c}{\bf Chit Chat}\\
\cmidrule(lr){1-2} \cmidrule(lr){3-5} \cmidrule(lr){6-8}
\bf RL  &  \bf Think    &   \bf Score   &   \bf Success   &   \bf Failure   &   \bf Score   &   \bf Success   &   \bf Failure \\
\midrule
\multicolumn{8}{c}{\bf \em Top-5 Models in Sentient Leaderboard}\\
\multicolumn{2}{r}{Gemini2.5-Pro-0605}          &   82.4    &   55\%  &   4\%   & 83.3 & 77\% & 11\% \\
\multicolumn{2}{r}{GPT-4o-0326}                 &   79.9    &   51\%  &   4\%   & 80.9 & 74\% & 17\%  \\
\multicolumn{2}{r}{GPT-4.1-0414}                &   68.2    &   35\%  &   13\%  & 77.1 & 65\% & 18\% \\
\multicolumn{2}{r}{Gemini2.5-Flash-Think-0520}  &   66.1    &   39\%  &   14\%  & 64.7 & 53\% & 27\% \\
\multicolumn{2}{r}{OpenAI-o3-0416}              &   62.7    &   32\%  &   14\%  & 83.0 & 66\% & 9\%  \\
\toprule
\multicolumn{8}{c}{\bf \em Our RLVER-Trained Models}\\
\multicolumn{2}{c}{Qwen2.5-7B-Instruct}        & 13.3    & 2\% & 76\% & 37.8 &27\% & 58\%\\
\midrule
\multirow{2}{*}{\bf PPO}     &   \bf \ding{56} &   61.7& 24\% & 23\%  &  53.4& 39\%& 37\%\\
                             &   \bf \ding{52} &    79.2& 42\% & 9\% & 62.1&52\%&30\%\\
\midrule
\multirow{2}{*}{\bf GRPO}    &   \bf \ding{56} &   68.3  & 26\% & 10\% & 49.2&34\%&40\%\\
                             &   \bf \ding{52} &    72.0 & 34\% & 10\%& 53.0& 45\%& 42\%\\
\bottomrule
\end{tabular}
\label{tab:sentient_leaderboard}
\end{table*}

Table~\ref{tab:sentient_leaderboard} presents the results of the proposed \method.

\paragraph{RLVER elevates a lightweight 7B model to near-frontier empathetic performance.}
The base model, Qwen2.5-7B-Instruct, struggles significantly on the Sentient Benchmark, scoring only 13.3 with a high failure rate (76\% of dialogues). In contrast, our RLVER-trained models demonstrate a remarkable improvement. Our best-performing model, trained with PPO and an explicit thinking step (``PPO + Thinking''), achieves a score of 79.2, representing a nearly six-fold increase over the base model. This result not only drastically increases the success rate from 2\% to 42\% but also brings our 7B model's performance in line with top-tier proprietary models like Gemini2.5-Pro (82.4), while substantially outperforming others such as Gemini2.5-Flash-Think (66.1) and OpenAI-o3 (62.7). This directly validates our primary contribution: the successful application of RL to enhance multi-turn empathetic dialogue capabilities in LLMs.

\paragraph{``Thinking'' models generally exhibit higher empathetic capabilities than ``non-thinking'' models after training.}
Experimental results show that models trained with a thinking scaffold consistently outperform their non-thinking counterparts on both the Sentient Benchmark and Chit-Chat tasks. When trained with PPO, the thinking model achieves a notable improvement from 61.7 to 79.2, surpassing the non-thinking variant. These results suggest that incorporating an explicit reasoning process may facilitate the emergence of higher-order empathetic strategies in LLMs. To further investigate this phenomenon, in \S\ref{sec:capability}, we present a detailed evaluation of the models’ empathetic behavior, demonstrating that eliciting reasoning enhances both the depth of empathy and the ability to identify users’ core concerns.

\paragraph{While both RL algorithms are effective, GRPO tends to offer greater training stability, whereas PPO may provide a higher performance ceiling.}
Our results also reveal a nuanced comparison between the PPO and GRPO algorithms. When training both thinking and non-thinking models, GRPO achieves stable improvements, reaching a score of 68.3 in the non-thinking setting and 72.0 in the thinking setting.
In contrast, PPO yields lower performance in the non-thinking case (61.7), but enables the thinking model to reach a higher performance ceiling (79.2). 
In \S\ref{sec:capability}, we further highlight an intriguing observation: PPO and GRPO induce different patterns in the development of model capabilities.

\begin{table*}[t]
\centering
\caption{Performance of our proposed methods on general tasks. We report results on the out-of-domain MATH500, LiveCodeBench \texttt{release\_v6} (Pass@1) code-generation, and IFEval benchmarks to assess instruction following performance.}
\begin{tabular}{cc ccc} 
\toprule
\multicolumn{2}{c}{\bf Model}   &   \multicolumn{3}{c}{\bf General Capability}\\
\cmidrule(lr){1-2} \cmidrule(lr){3-5}
\bf RL  &  \bf Think    &   \bf Math500    &   \bf LiveCodeBench    &   \bf IFEval\\
\midrule
\multicolumn{2}{c}{Qwen2.5-7B-Instruct}        & 77.8 & 26.7 &70.4 \\
\midrule
\multirow{2}{*}{\bf PPO}     &   \bf \ding{56} & 76.2 &33.3  &72.3 \\
                             &   \bf \ding{52} & 76.6 &28.0  &68.6 \\
\midrule
\multirow{2}{*}{\bf GRPO}    &   \bf \ding{56} & 77.4 & 30.0 &72.1\\
                             &   \bf \ding{52} & 75.2 &29.3  & 69.7\\
\bottomrule
\end{tabular}
\label{tab:general_tasks}
\end{table*}

\paragraph{Specialization in empathetic reasoning is achieved with minimal impact on general capabilities.}
A critical aspect of fine-tuning is ensuring that specialization in one domain does not lead to catastrophic forgetting in others. Our evaluation on out-of-domain benchmarks in Table~\ref{tab:general_tasks} shows that our training successfully avoids this pitfall. While there is a minor decrease in mathematical reasoning performance on Math500 (from 77.8 to 76.6 for our best PPO model), performance on the LiveCodeBench code-generation benchmark is maintained or even improved (from 26.7 to 28.0). Moreover, the model’s ability to follow instructions, as measured by IFEval, remains stable (from 70.4 to 68.6). This demonstrates that our framework can cultivate sophisticated emotional intelligence while preserving the model's core general-purpose functionalities, making it a practical and well-rounded solution.

\subsection{Qualitative Analysis of Trained Agents}\label{sec:capability}

\begin{figure}[h]
    \centering
    \subfloat[Models trained with PPO]{\includegraphics[width=0.48\linewidth]{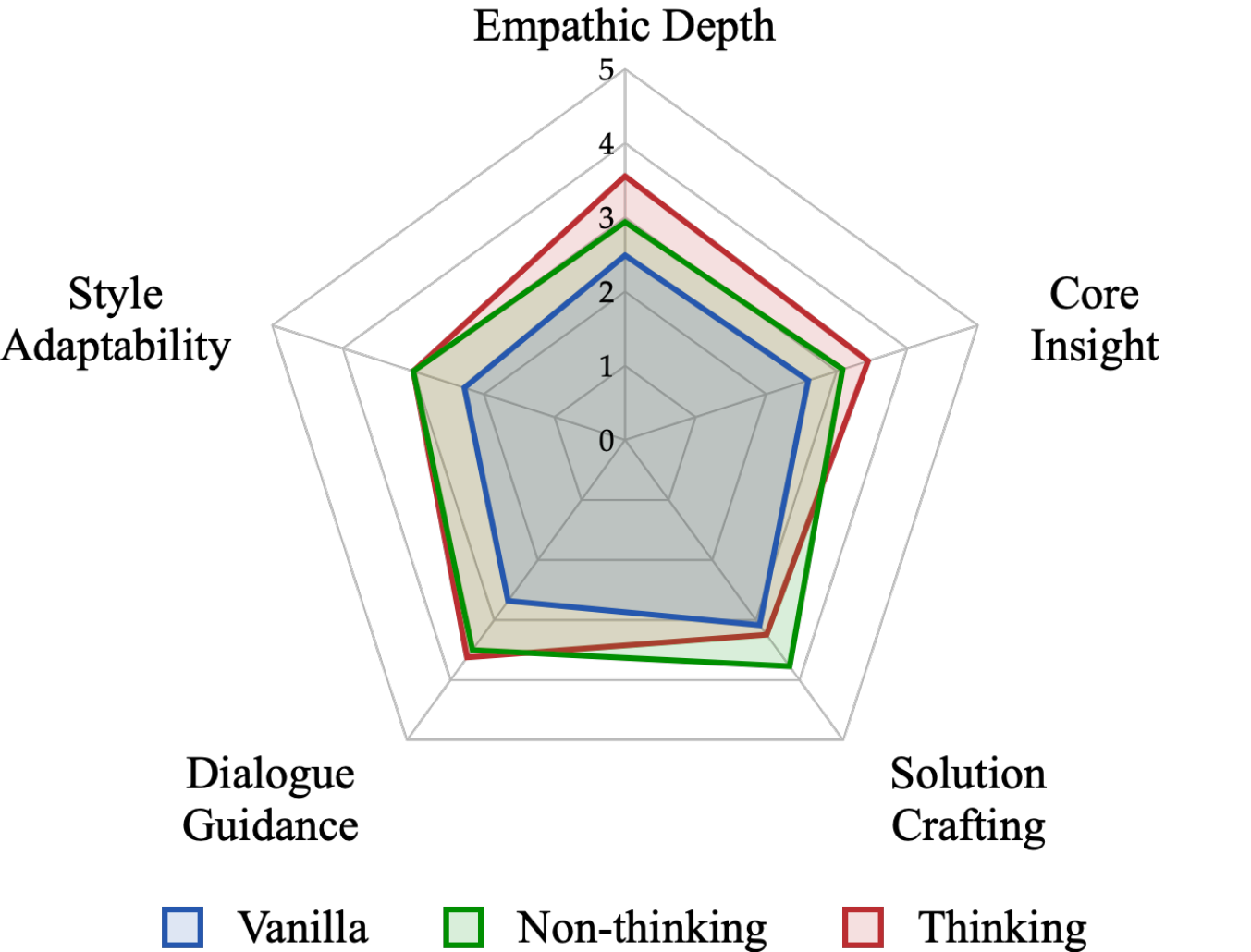}} \hspace{0.02\linewidth}
    \subfloat[Models trained with GRPO]{\includegraphics[width=0.48\linewidth]{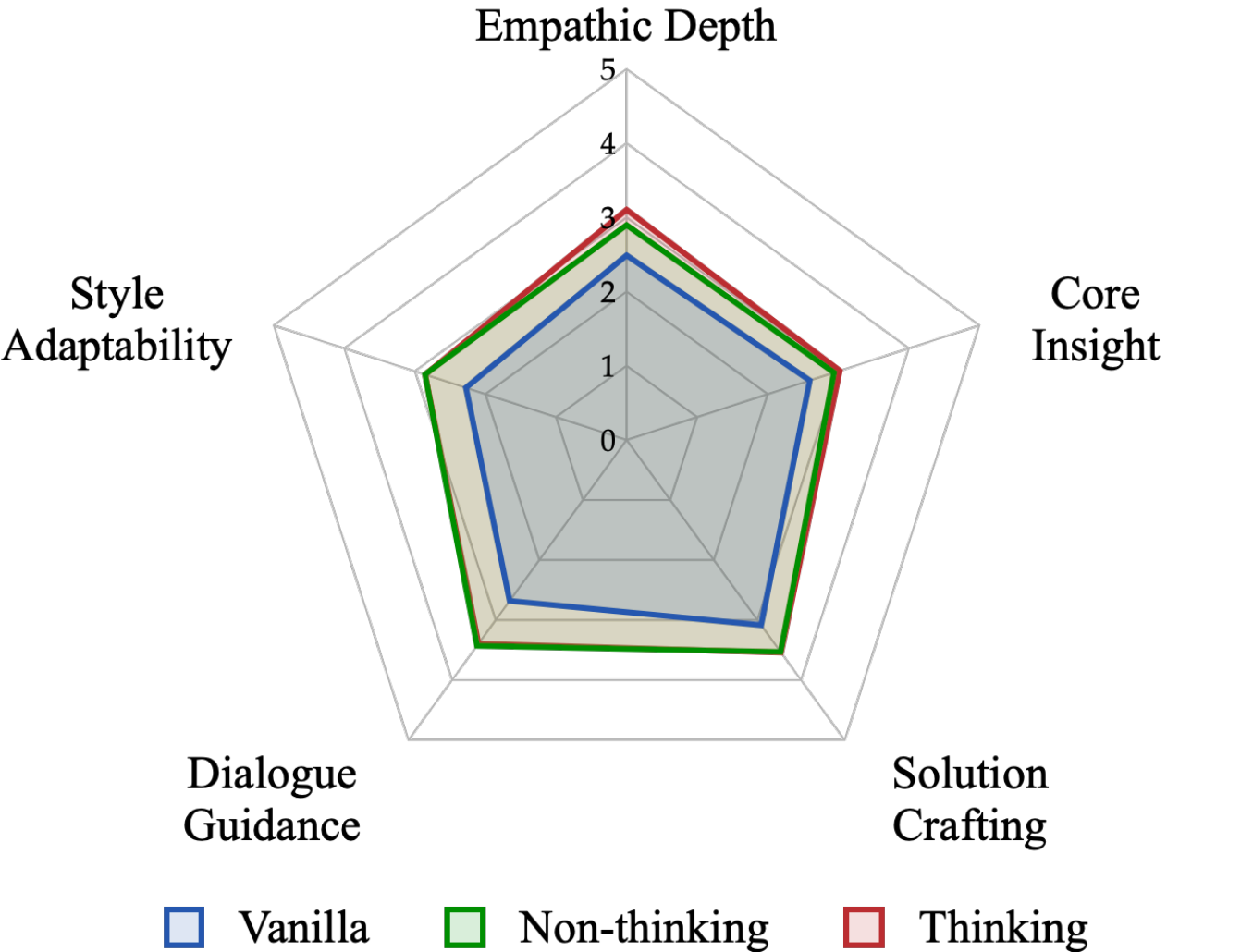}} \\\vspace{0.02\linewidth}
    \caption{Qualitative analysis of five core capabilities of the trained models.}
    \label{fig:capability_analysis}
\end{figure}

In order to further investigate the models’ capability improvements after RL training, we formalize a comprehensive evaluation framework encompassing five core competencies in the empathetic dialogue task:

\begin{itemize}[leftmargin=12pt]
\item \textit{\textbf{Empathic Depth}}: the model’s ability to move beyond templated responses to genuinely identify and comprehend the user’s complex, deep-seated emotions, and to accurately validate those emotions using precise, warm, and emotionally resonant language.
\item \textit{\textbf{Core Insight}}: the model’s ability to integrate and distill information from the user’s fragmented narrative to construct a holistic understanding of their situation ultimately identifying the user’s unmet emotional needs.
\item \textit{\textbf{Solution Crafting}}: the model’s ability to deliver actionable, personalized, and empowering suggestions. This entails not merely offering an answer, but proposing a step-by-step course of action that enables the user to feel genuinely capable of carrying it out.
\item \textit{\textbf{Style Adaptability}}: the model’s ability to flexibly adjust its communicative role and linguistic style in response to the conversational context, the user’s implicit preferences, and the long-term relational dynamic.
\item \textit{\textbf{Dialogue Guidance}}: the model’s ability to proactively and flexibly guide the conversation—based on the user’s emotional state—from emotional expression toward constructive problem-solving, while staying aligned with the user’s pace and needs.
\end{itemize}

We used LLM-as-a-Judge for evaluating the five core capabilities, and the results are shown in Figure \ref{fig:capability_analysis}. More experiment details can be found in Appendix \S \ref{app:capability}.

\paragraph{RLVER brings consistent improvement across five core capabilities.}
As shown in Figure~\ref{fig:capability_analysis}, models trained with RLVER—regardless of the specific training strategy—consistently outperform the base model across all five core dimensions. By quantilizing the assessment of these capabilities, we not only gain deeper insight into the behavioral differences induced by different strategies, but also provide an external and objective evaluation—beyond the test set—that supports the effectiveness of our training framework in enhancing key empathetic abilities.

\paragraph{Thinking models tend to excel in empathy and insight, while non-thinking models may specialize in action.}
With PPO training, we observe a clear divergence in the capability profiles of thinking and non-thinking models. The thinking model exhibits marked improvements in Core Insight (3.44) and Empathic Depth (3.56), demonstrating a strong ability to identify core user needs and to recognize deep emotions through precise, validating responses.
In contrast, the non-thinking model shows greater gains in Solution Crafting (3.77), emphasizing actionable, context-aware support through concrete suggestions or behavioral prompts. This pattern suggests that the thinking model benefits from explicit reasoning prior to response generation, enabling it to better infer the user’s emotional state and underlying concerns. The non-thinking model, lacking such reasoning, appears to compensate by offering more tangible and personalized solutions to assist the user.

\paragraph{PPO promotes higher ceilings in specific capabilities, while GRPO supports more balanced and stable development.}
A comparison between PPO and GRPO training reveals that GRPO facilitates more balanced and stable improvements across all five capabilities, while PPO tends to amplify specific strengths depending on the training strategy. The thinking model trained with PPO reaches higher performance ceilings in Core Insight (3.44 vs. 3.02 under GRPO) and Empathic Depth (3.56 vs. 3.10), while the non-thinking model trained with PPO achieves a higher ceiling in Solution Crafting (3.77 vs. 3.53). Aligned with the strong performance of the PPO-thinking model in Table~\ref{tab:sentient_leaderboard}, these findings suggest that in empathetic dialogue tasks, once a baseline level of competence is achieved across all dimensions (e.g., around 3.0), selectively enhancing high-impact abilities—such as Core Insight and Empathic Depth—may lead to greater practical effectiveness.

\subsection{Impact of Training Environment and Reward}
In RLVER, a key idea is to use self-consistent and scalable user simulators as training environments, with their emotional changes serving as reward signals. As a result, the outcomes of training are closely influenced by the behavioral characteristics of the user simulators. In previous sections, we have demonstrated the effectiveness of using these simulators as both environment and reward sources. In this section, we further investigate how variations in user simulator behavior impact the training outcomes.

Specifically, we compare the vanilla user simulator (vanilla version) with a more challenging variant (challenging version)—one that imposes stricter demands on the dialogue model and is more reserved in expressing its thoughts and emotions. \textbf{Intuitively, the challenging version requires the model to exhibit stronger general capabilities}, including more effectively identifying the user’s unmet emotional needs, demonstrating deeper empathy, and showing greater strategic flexibility and dialogue guidance.
We describe the behavioral characteristics of the two user simulator variants using two metrics: \textbf{\textit{Strategy Acceptance Rate}} and \textbf{\textit{Emotion and Need Expression Level}}. The detailed quantilized feature values are presented in Table~\ref{tab:setting_comparison} (detailed construction of the challenging version and the metric evaluation can be found in Appendix \ref{app:setting_comparison_detail}).

\begin{table}[ht]
\centering
\caption{Comparison of vanilla and challenging user simulator construction.}
\resizebox{\textwidth}{!}{
\begin{tabular}{lcccc}
\toprule
\multirow{2}{*}{Metric} & \multicolumn{2}{c}{Vanilla user simulator} & \multicolumn{2}{c}{Challenging user simulator} \\
\cmidrule(lr){2-3} \cmidrule(lr){4-5}
 & Non-thinking& Thinking & Non-thinking & Thinking \\
\midrule
Sentient Benchmark  & 61.7 & 79.2 & 19.8 & 66.4 \\
Sentient Benchmark (challenging) & 47.7 & 59.6 & 25.9 & 44.7 \\
\midrule
Strategy acceptance rate & \multicolumn{2}{c}{52.4\%} & \multicolumn{2}{c}{33.1\%} \\
Emotion and need expression level & \multicolumn{2}{c}{78.6\%} & \multicolumn{2}{c}{63.6\%} \\
\bottomrule
\end{tabular}
}
\label{tab:setting_comparison}
\end{table}

\begin{figure}[h]
    \centering
    \subfloat[PPO-thinking models]{\includegraphics[width=0.48\linewidth]{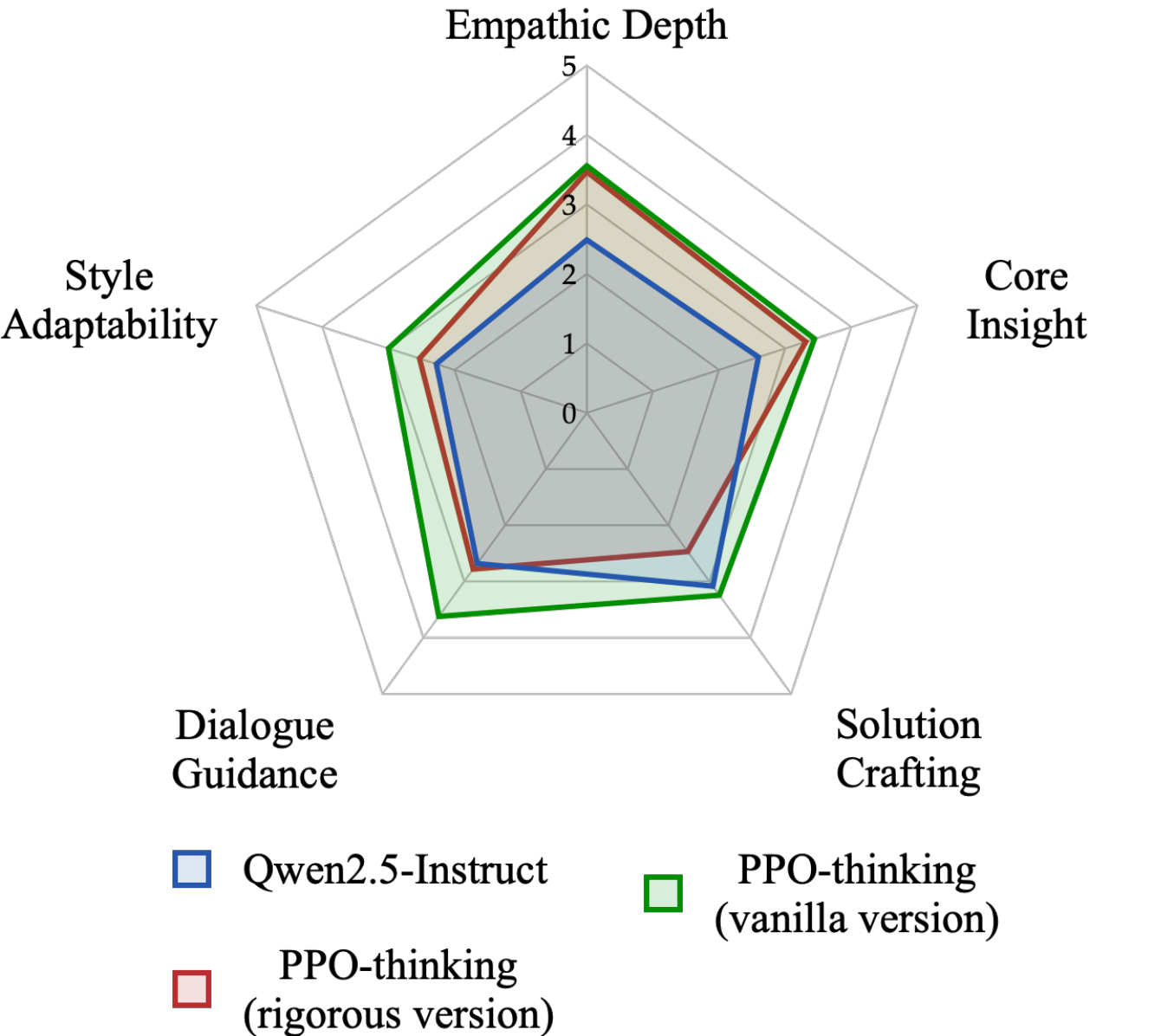}} \hspace{0.02\linewidth}
    \subfloat[PPO-non-thinking models]{\includegraphics[width=0.48\linewidth]{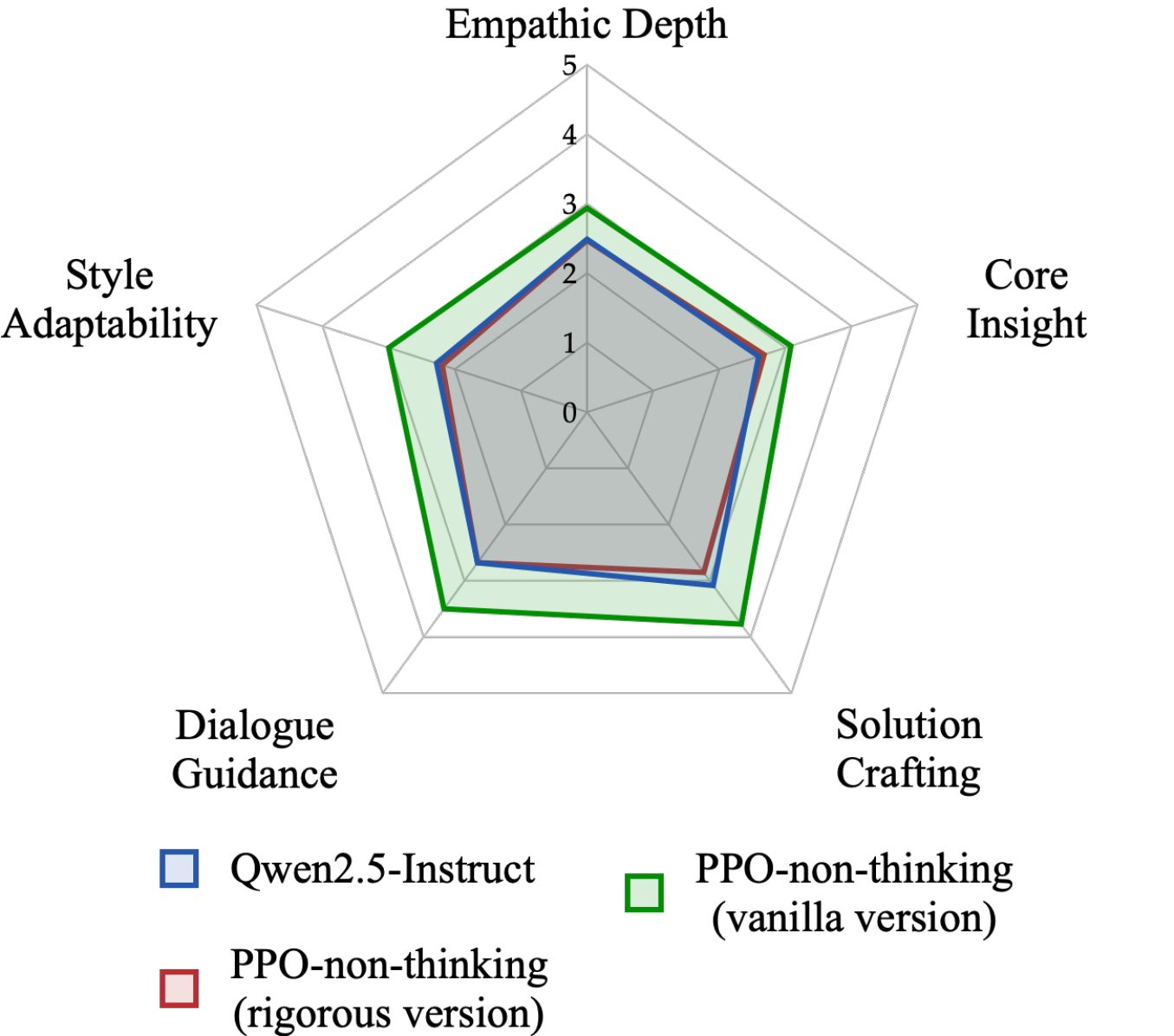}} \\\vspace{0.02\linewidth}
    \caption{Qualitative analysis of training outcomes with vanilla and challenging user simulators.}
    \label{fig:setting_capability_analysis}
\end{figure}

\paragraph{More challenging environments and reward modeling do not necessarily yield better outcomes} In Table~\ref{tab:setting_comparison}, we present the test results of models trained with the vanilla and challenging user simulators, both using PPO. Notably, models trained with the challenging simulator consistently underperform compared to their vanilla-trained counterparts. When evaluated on the Sentient Benchmark, the thinking model trained with the challenging simulator scores 66.4, notably lower than the 79.2 achieved with the vanilla simulator. The non-thinking model performs even worse, dropping from 61.7 (vanilla) to just 19.8. To extend the comparison, we instantiate the Sentient Benchmark with the challenging user simulator, resulting in a challenging version of the original benchmark. However, even on this harder benchmark, models trained with the challenging simulator still perform worse than those trained with the vanilla version (thinking model: 59.6 vs. 44.7; non-thinking model: 47.7 vs. 25.9).

These results suggest that in settings where user simulators serve as both the environment and reward signal, more challenging configurations do not necessarily lead to better learning outcomes. On the contrary, moderately demanding and well-calibrated simulators may result in more effective training. A possible explanation is that overly strict or reserved simulators restrict feedback during the model's exploration phase, making it difficult for models—especially those with limited initial capabilities—to discover useful growth trajectories. In contrast, user simulators with more moderate requirements may provide richer feedback, facilitating more diverse strategy exploration and enabling more comprehensive skill development throughout training.

\paragraph{Thinking models exhibit greater robustness to environment variations than non-thinking models.} 
As shown in Table~\ref{tab:setting_comparison}, thinking models maintain relatively strong performance even under the challenging setting, with scores dropping from 79.2 to 66.4. In contrast, non-thinking models experience a dramatic performance decline, falling from 61.7 to just 19.8 when the environment changes. In Figure~\ref{fig:setting_capability_analysis}, we further analyze the development of core capabilities after training. We find that non-thinking models show little to no improvement over the original model across all capabilities under the challenging setting. In contrast, thinking models still exhibit clear gains in Empathic Depth, Core Insight, and Style Adaptability, demonstrating their resilience and adaptability even under more demanding training conditions. 

Interestingly, \textbf{the development pattern of the thinking model’s capabilities appears to align closely with the behavioral characteristics of the challenging simulator}. As the simulator demands deep empathetic reasoning but offers limited explicit feedback, the model shows targeted gains in Core Insight and Empathic Depth. Specifically, its notable gains in Core Insight indicate an enhanced ability to infer unspoken emotional needs and intentions, while improvements in Empathic Depth reflect increased sensitivity to subtle emotional cues and a greater capacity for emotional validation—both of which are essential for eliciting reward under sparse feedback conditions.
Conversely, the model’s Solution Crafting ability declines, likely because most rollouts never reach the stage where actionable suggestions are appropriate—typically after establishing empathy. Likewise, gains in Dialogue Guidance and Style Adaptability are limited, as the simulator offers few opportunities to explore dynamic strategies or role flexibility under restrictive, uncooperative conditions.

\subsection{Learning Curves of Emotion Scores}

\begin{figure}[t]
    \centering
    \subfloat[Emotion@PPO]{\includegraphics[width=0.45\linewidth]{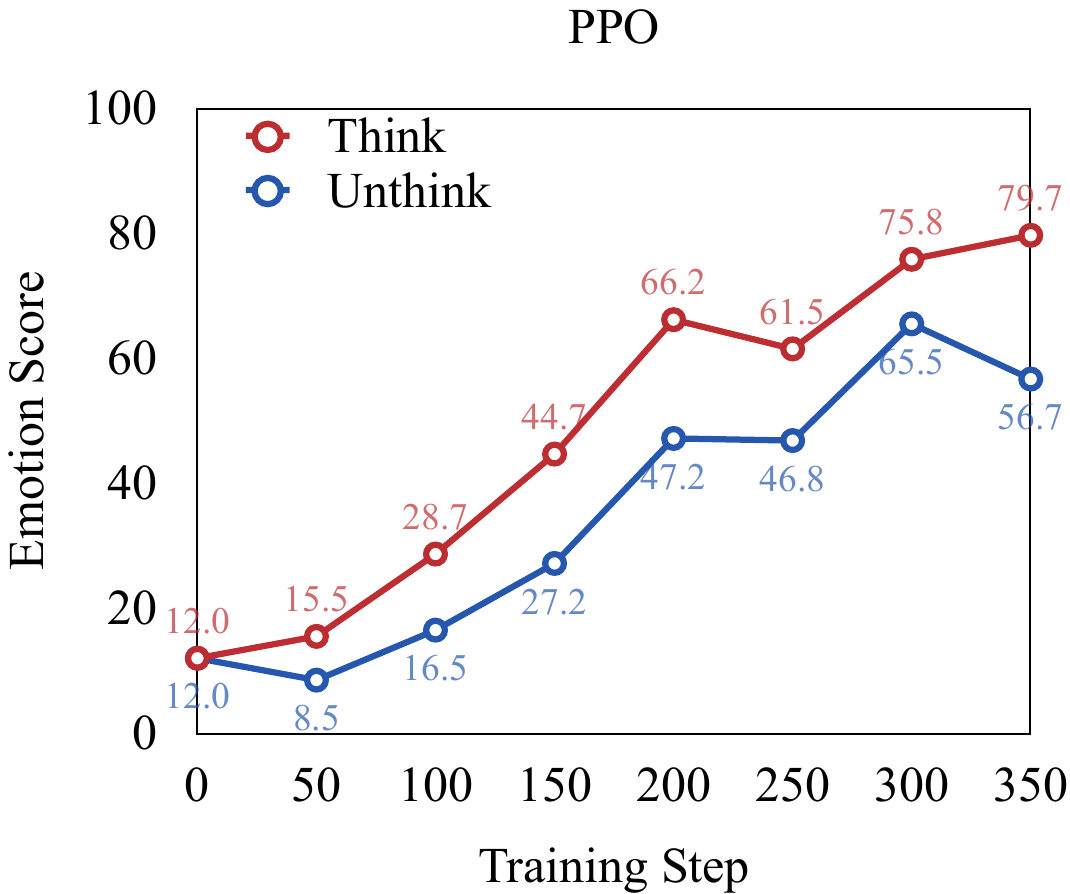}} \hspace{0.05\linewidth}
    \subfloat[Emotion@GRPO]{\includegraphics[width=0.45\linewidth]{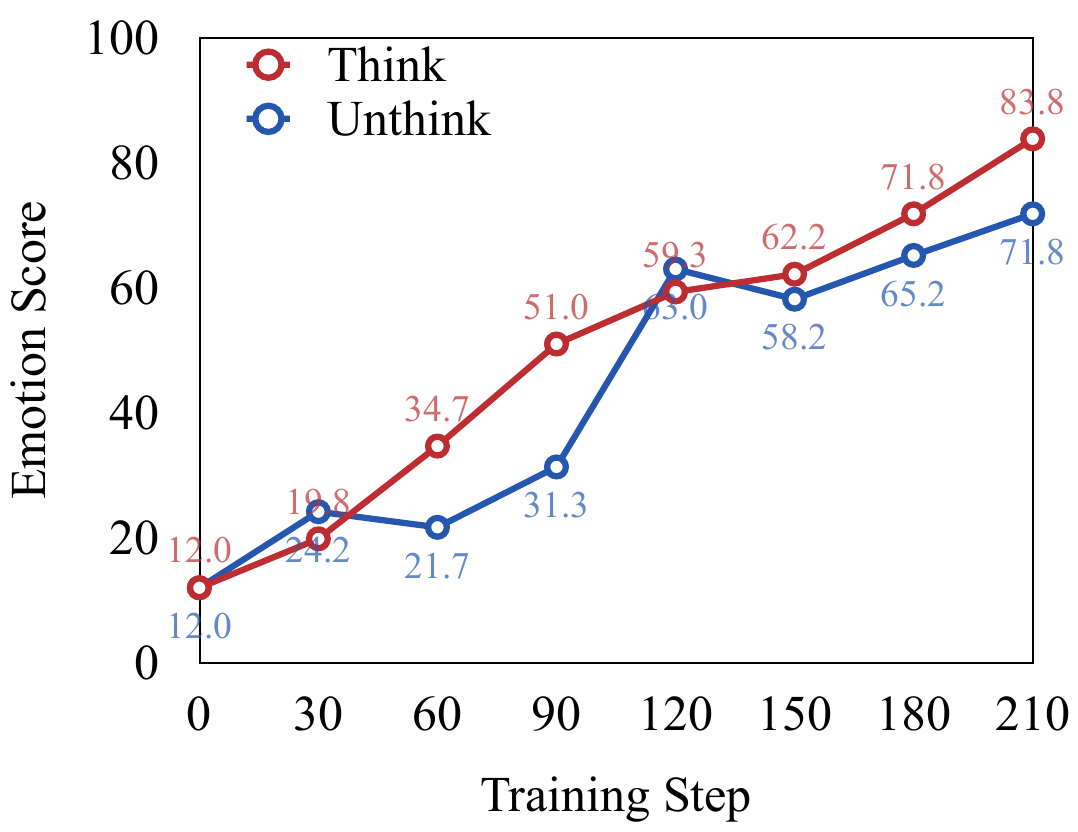}} \\\vspace{0.02\linewidth}
    \subfloat[Tokens@PPO]{\includegraphics[width=0.45\linewidth]{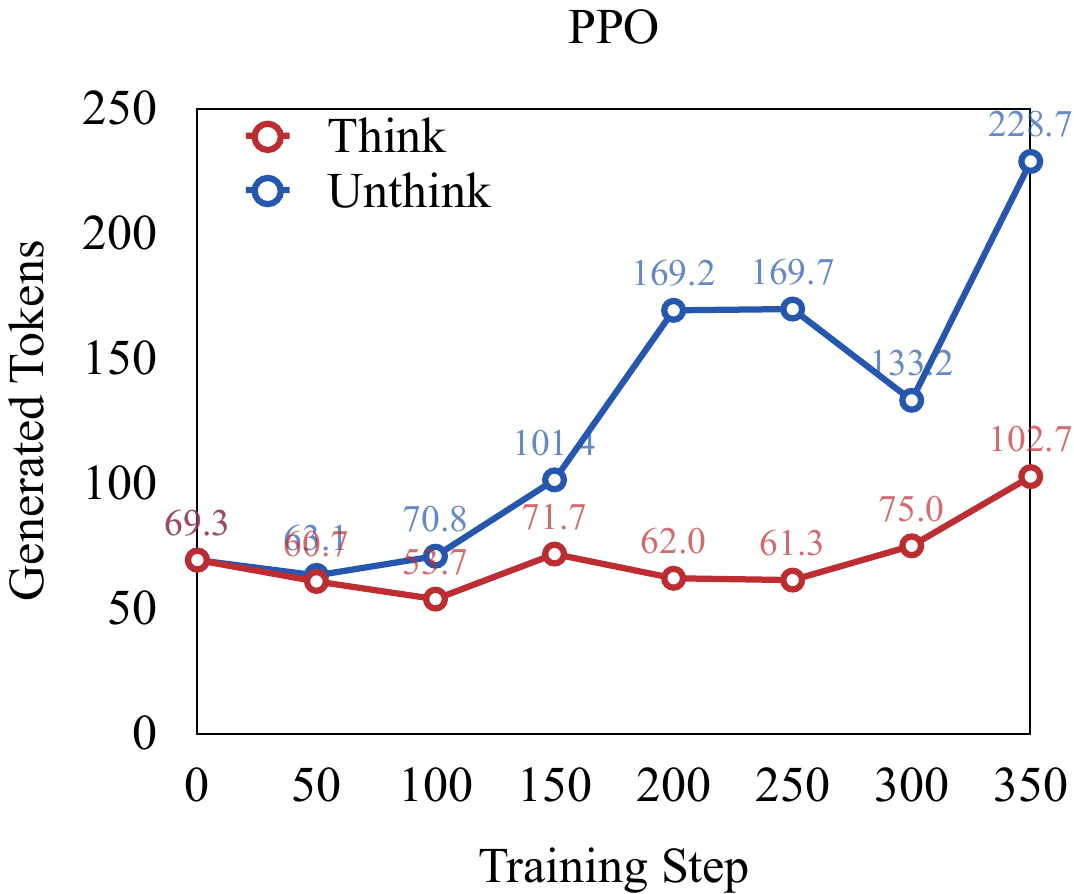}} \hspace{0.05\linewidth}
    \subfloat[Tokens@GRPO]{\includegraphics[width=0.45\linewidth]{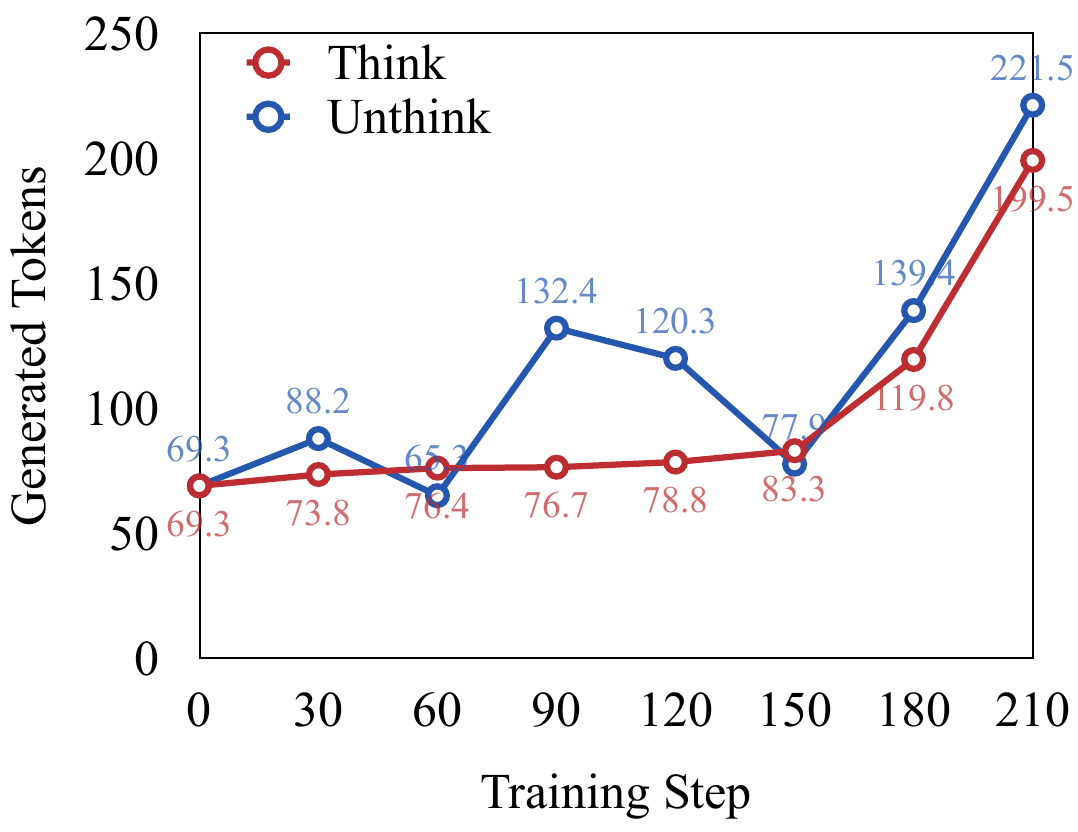}}
    \caption{Learning curves for (a,\,b) emotion scores and (c,\,d) generated token counts.}
    \label{fig:curves_emotion}
\end{figure}

In this section, we analyze the learning curves of our approach with respect to emotion scores and generated tokens. We randomly sample 30 instances from the test set and report the corresponding emotion scores and generated token counts produced by the models throughout training. Figure~\ref{fig:curves_emotion} presents these results.

\paragraph{The "think-then-say" scaffold is an important contributor to performance and stability.}
Across both optimization algorithms, inserting an explicit reasoning step is the most influential intervention. As shown in Figure~\ref{fig:curves_emotion}(a)–(b), scaffolded models learn faster and attain markedly higher emotion scores. Under PPO, the scaffold averts the catastrophic collapse observed in the baseline (79.7 vs.\,56.7). Under GRPO, it raises an already stable learner to the highest score recorded (83.8). These findings substantiate Contribution 2: the scaffold simultaneously accelerates and stabilizes learning.

\paragraph{The RLVER framework is robust across policy-optimization algorithms.}
Framework effectiveness does not hinge on a particular optimizer. The reasoning scaffold propels GRPO to the overall peak (83.8) while acting as a crucial regularizer for PPO, converting an erratic trajectory into a consistently successful one. This dual achievement reinforces Contribution 1, demonstrating that RLVER is general-purpose rather than algorithm-specific.

\paragraph{Empathetic skill is learned strategically, not via verbose reward hacking.}
Figures~\ref{fig:curves_emotion}(c)–(d) confirm that superior emotion scores are not merely a by-product of generating longer texts. The PPO-Think model is initially more concise than its baseline, with token counts rising only after empathetic dominance is established. The GRPO-Think model remains less verbose than its counterpart for most of training. These trends refute the verbosity-as-shortcut hypothesis and support the claim that the model develops a genuinely empathetic style.

In summary, verifiable emotion rewards coupled with a reasoning scaffold provide a reliable path to empathy.
The synergy between verifiable rewards and the ``think-then-say'' structure consistently steers a 7B model toward elite empathetic performance. Its resilience across optimizers, resistance to reward hacking, and pronounced impact on learning stability and efficiency confirm RLVER as a practical, robust methodology for building emotionally intelligent agents.

\subsection{Learning Curves of Empathetic Strategies}

\begin{table*}[t]
    \centering
    \begin{tabular}{ll}
        \toprule
         \bf Group &\bf Strategy \\
         \midrule
        \multirow{3}{*}{(A) \bf Praise} & (A-1) Praising the client's qualities\\
        & (A-2) Praising the client's positive thoughts \\
        & (A-3) Praising the client's actions \\
        \midrule
        \multirow{4}{*}{\shortstack{(B) \bf Deep Empathy}}& (B-1) Providing empathy via restating the client's problem \\
         & (B-2) Deeper empathy to understand the client's hidden intention \\
         & (B-3) Self-disclosure that provides agreement with the client's view \\
         & (B-4) Self-disclosure that introduces the supporter's own story \\
        \midrule
        \multirow{2}{*}{(C) \bf Emotional Venting}  & (C-1) Expressing willingness to hear the client's thoughts \\
        & (C-2) Helping the client to vent negative feelings \\
        \midrule
        \multirow{1}{*}{\shortstack{(D) \bf Advice Provision}}& (D-1) Advice specific to the client's situation \\
        \midrule
        \multirow{1}{*}{\shortstack{(E) \bf Problem Analysis}}& (E-1) Analysis of the client's issue \\
        \bottomrule
    \end{tabular}
    \caption{Details of the support strategy categorization.} 
    \label{tab:strategy_category}
\end{table*}

To distinguish the response behaviors of trained models, we classify each model reply according to a predefined list of support strategies. Following \cite{liu2021towards}, we group these strategies into the five categories shown in Table~\ref{tab:strategy_category}.

To analyze strategy usage, we prompt DeepSeek-V3 to act as a judge: for each dialogue turn, it identifies all strategies present in the model's output. We then aggregate these annotations across the conversation and report the proportion of turns in which each strategy appears.

\begin{figure}[t]
    \centering
    \subfloat[PPO-non-thinking]{\includegraphics[width=0.45\linewidth]{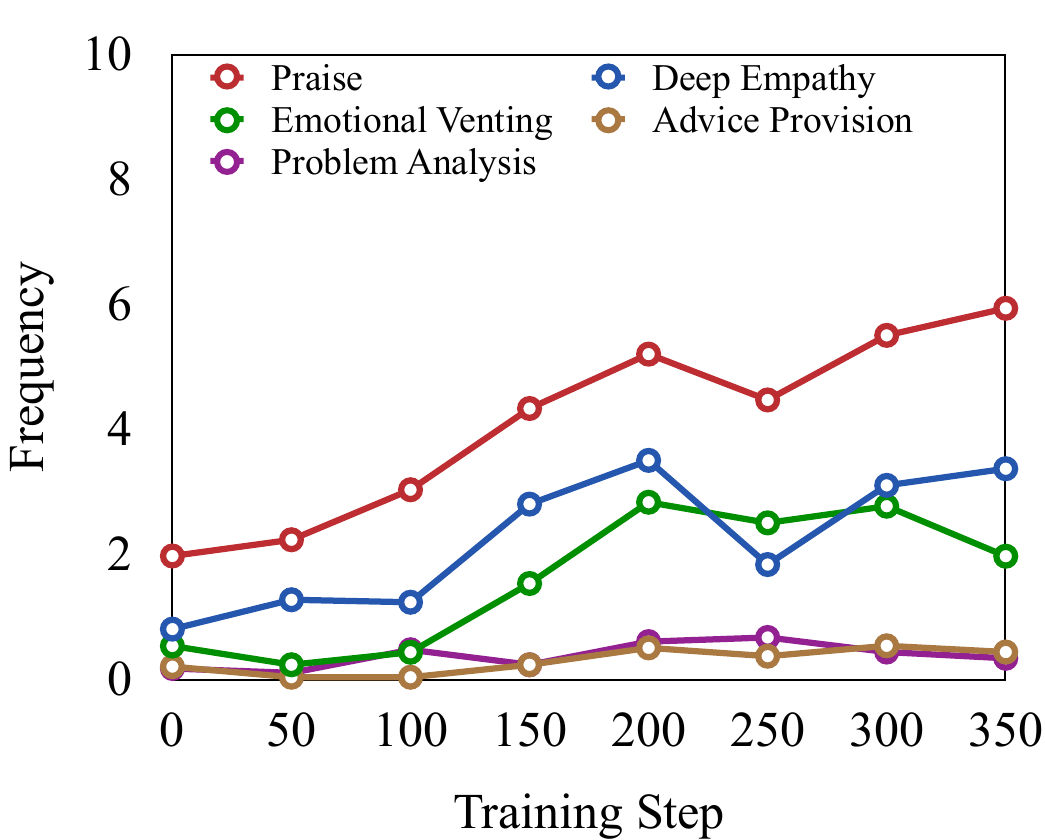}} \hspace{0.05\linewidth}
    \subfloat[PPO-thinking]{\includegraphics[width=0.45\linewidth]{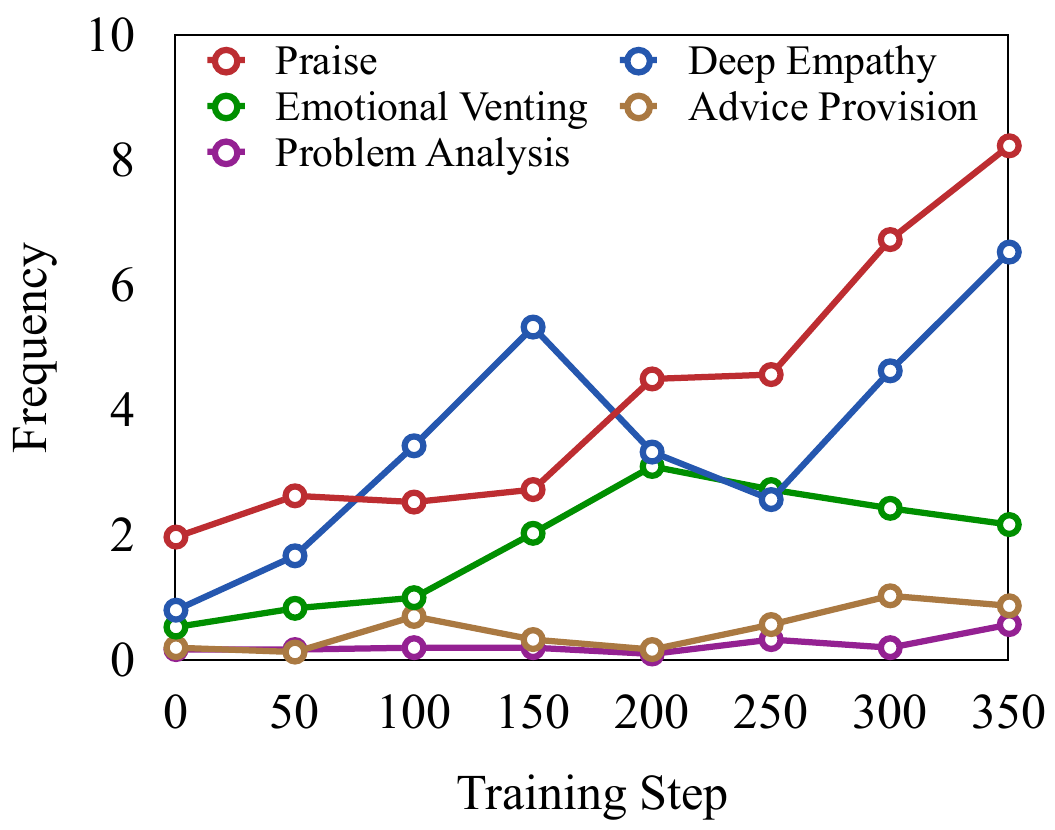}} \\\vspace{0.02\linewidth}
    \subfloat[GRPO-non-thinking]{\includegraphics[width=0.45\linewidth]{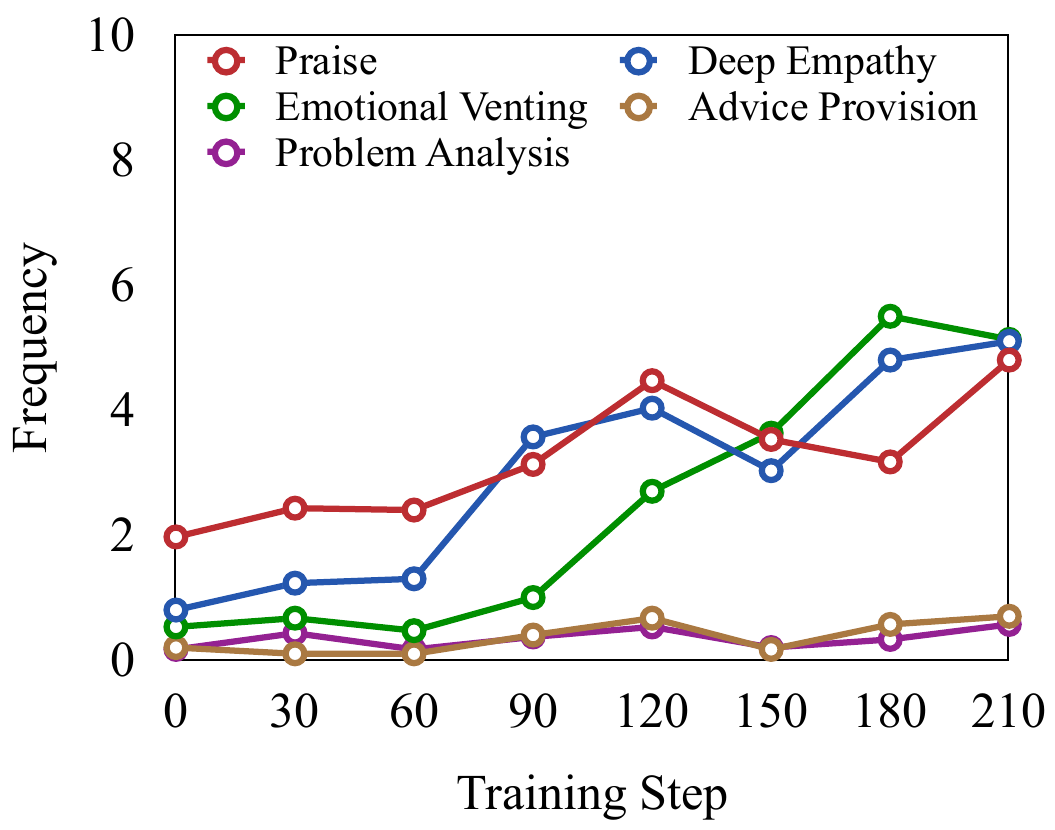}} \hspace{0.05\linewidth}
    \subfloat[GRPO-thinking]{\includegraphics[width=0.45\linewidth]{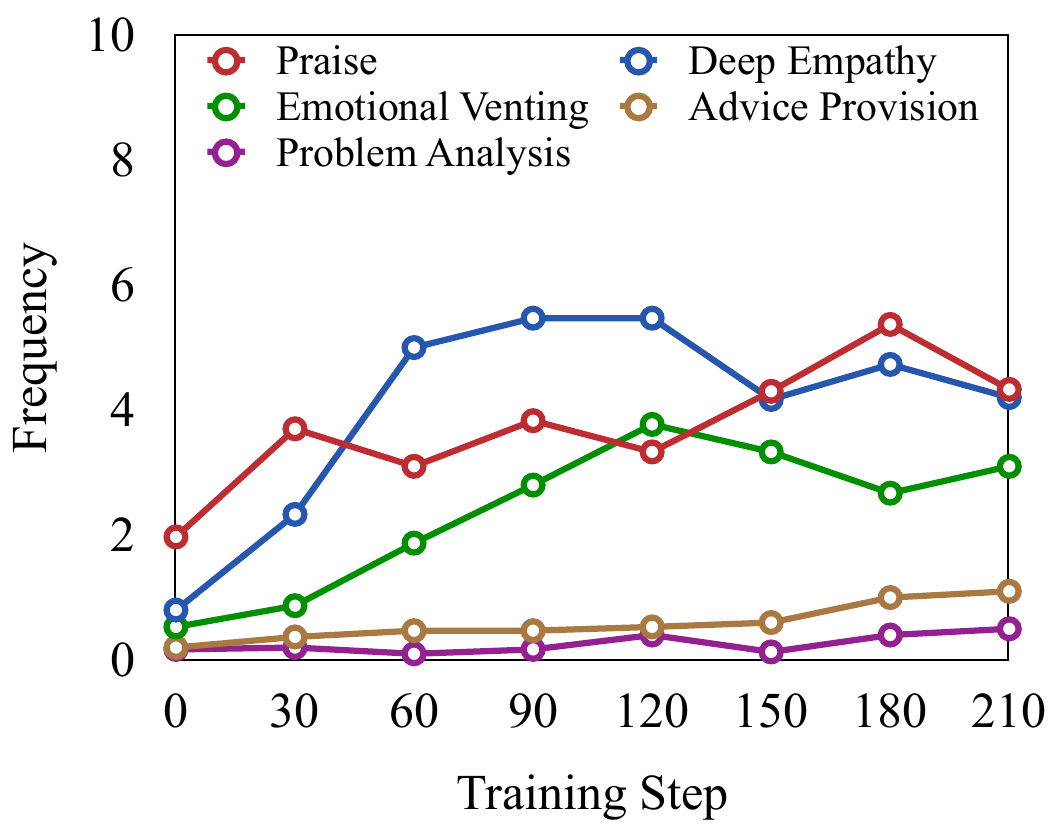}}
    \caption{Frequency of empathetic strategies during the training.}
    \label{fig:curves_strategies_frequency}
\end{figure}

\begin{figure}[t]
    \centering
    \subfloat[PPO-non-thinking]{\includegraphics[width=0.45\linewidth]{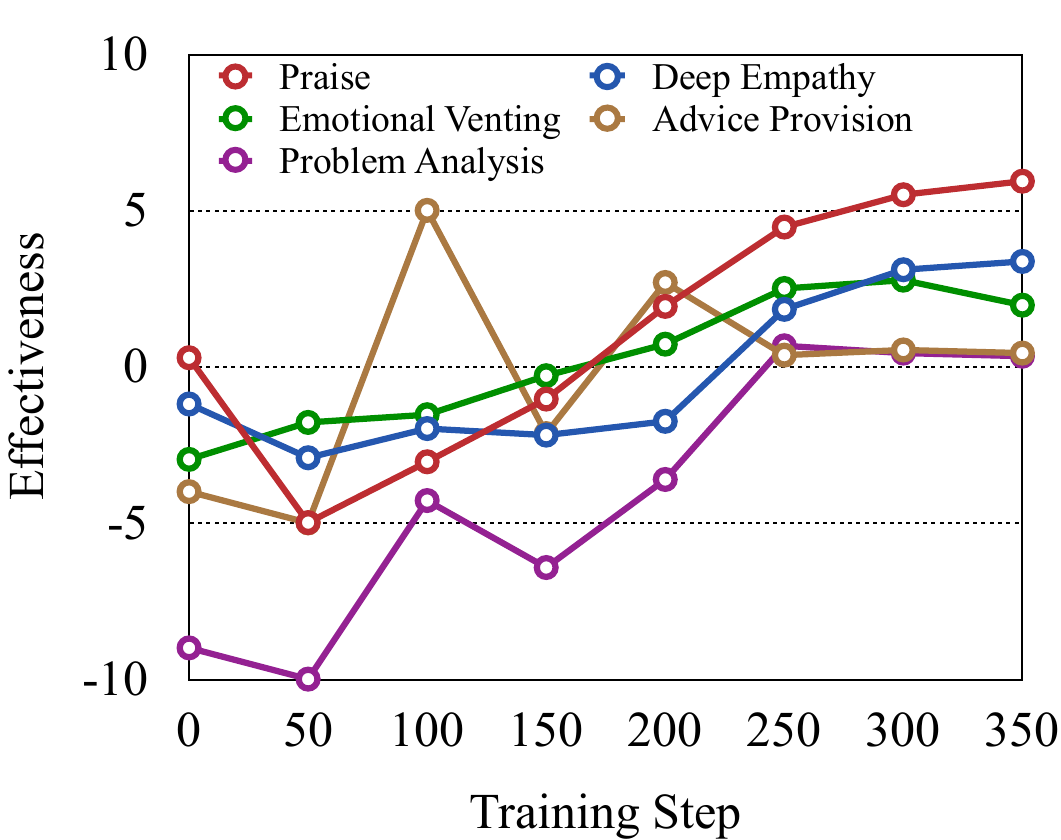}} \hspace{0.05\linewidth}
    \subfloat[PPO-thinking]{\includegraphics[width=0.45\linewidth]{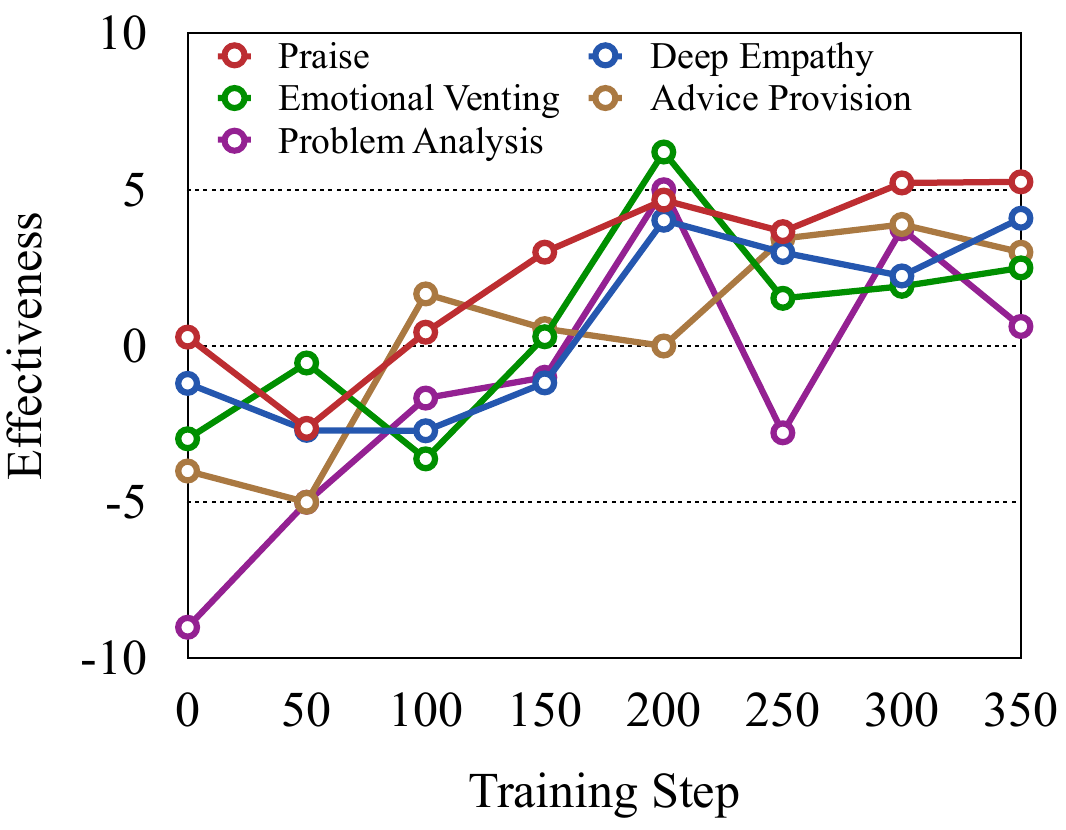}} \\\vspace{0.02\linewidth}
    \subfloat[GRPO-non-thinking]{\includegraphics[width=0.45\linewidth]{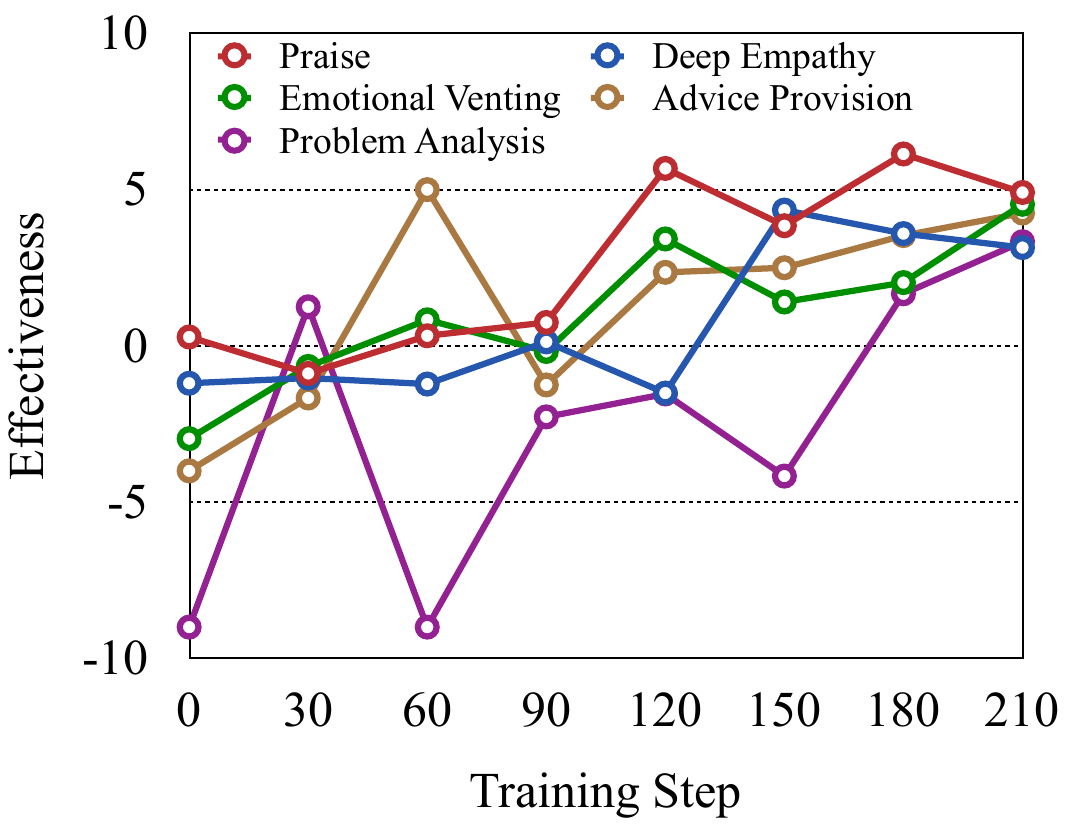}} \hspace{0.05\linewidth}
    \subfloat[GRPO-thinking]{\includegraphics[width=0.45\linewidth]{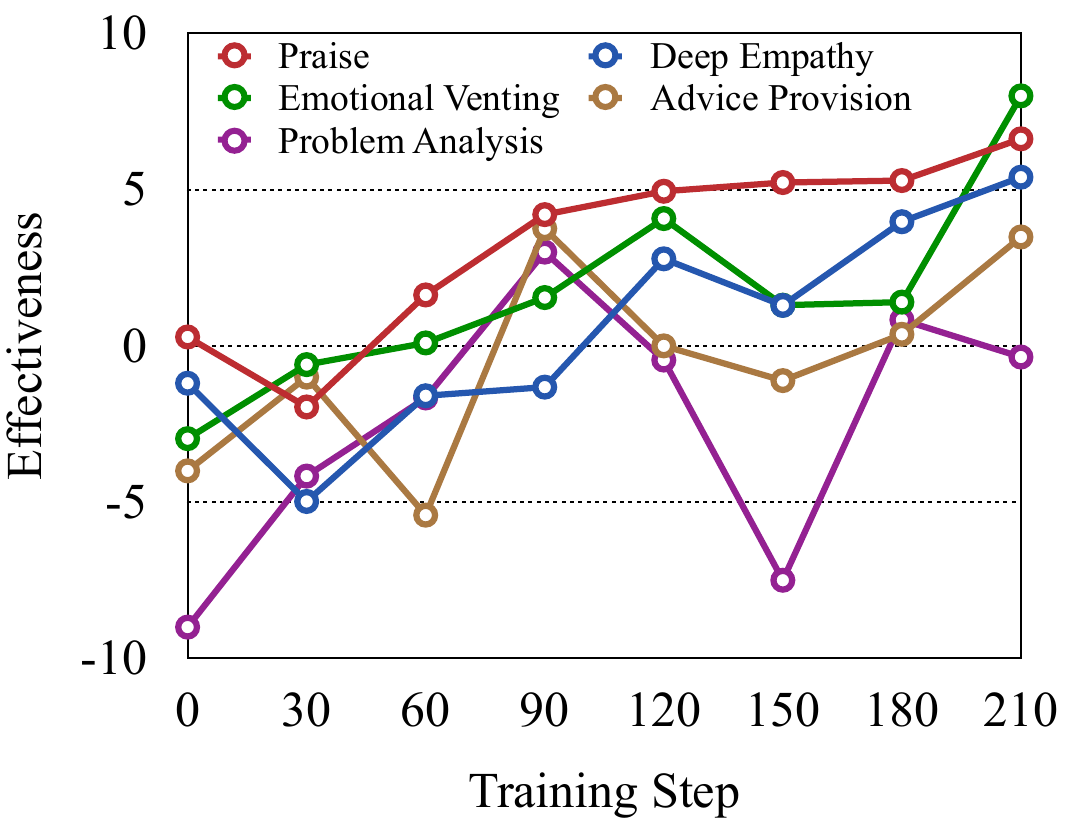}}
    \caption{Contribution of empathetic strategies during the training.}
    \label{fig:curves_strategies_contribution}
\end{figure}

Beyond raw frequency, we also examine how each strategy contributes to emotional improvement. To evaluate the appropriateness and effectiveness of strategy usage, we define the \emph{Strategy Contribution} (SC) for each strategy type as follows:
$$
\text{SC}= \frac{1}{N}\sum_{i=1}^{N}\  \text{EmoChange}(s_i)
$$
where the sum is over the $N$ instances where strategy $s$ was used. Figures~\ref{fig:curves_strategies_frequency} and~\ref{fig:curves_strategies_contribution} show the results.

\paragraph{Verifiable emotion rewards successfully steer the agent from shallow solutions to genuine empathy.}
At the start of training, nearly all strategies—including empathetic ones—yield a negative emotional contribution (Figure~\ref{fig:curves_strategies_contribution}), indicating that the base model lacks true empathetic skill. The RLVER framework promptly corrects this deficit. The agent learns to favor strategies that consistently improve the verifiable emotion score, causing the frequency of ``Praise'' and ``Deep Empathy'' to rise markedly while their contributions shift from negative to strongly positive. In contrast, the frequencies of ``Advice Provision'' and ``Problem Analysis'' remain low and their contributions volatile, because the deterministic reward signal prevents the agent from exploiting these low-effort shortcuts. This directly validates our first contribution: RLVER provides a robust learning signal for acquiring authentic empathy.

\paragraph{The think-then-say'' scaffold appears to support the emergence and stabilization of higher-order empathetic strategies.} A comparison between models trained with and without the scaffold highlights the potential role of explicit reasoning. The \textsc{PPO-thinking} model (Figure~\ref{fig:curves_strategies_frequency}(b)) shows a steady increase in the use of Deep Empathy''—rising more than 8$\times$ (from 0.8 to 6.53)—along with a notable rise in its emotional contribution ($+4.09$). In contrast, the \textsc{PPO-non-thinking} model (Figure~\ref{fig:curves_strategies_frequency}(a)) achieves a lower final frequency (2.1) and displays less stable progression. Notably, non-thinking models tend to exhibit late-stage declines in key strategies, suggesting potential instability. These observations suggest that the reasoning scaffold may play an important role in supporting the learning and retention of complex empathetic strategies, especially in longer training horizons.

\paragraph{PPO may better leverage the reasoning scaffold for effective policy optimization.}
While both algorithms benefit from the scaffold, \textsc{PPO-thinking} shows a relatively stronger and more stable learning trajectory. As shown in Figs.\ref{fig:curves_strategies_frequency} (b) and (d), PPO displays a smoother increase in using key strategies such as “Praise” and “Deep Empathy”. GRPO, though also effective, shows greater variance—occasionally producing sharp gains (e.g., in “Emotional Venting”) but with less consistency overall. Furthermore, the emotional contributions associated with \textsc{PPO-thinking} (Figure\ref{fig:curves_strategies_contribution}(b)) tend to reach higher positive values for core empathetic strategies. These observations suggest that PPO may be more effective in leveraging the structure provided by explicit reasoning, though GRPO could still hold advantages in specific settings, particularly those prioritizing stability or safety.

\paragraph{The framework learns nuanced strategy application, not just increased frequency.}
A key finding is the qualitative improvement in how strategies are employed. ``Advice Provision'' remains infrequent across all training runs ($<1.1$), yet its contribution transforms from strongly negative ($-4.0$) to positive values in \textsc{PPO-thinking}, demonstrating that the agent learns \emph{when} and \emph{how} to offer advice appropriately. Similarly, ``Emotional Venting'' becomes both more frequent and more effective, showing the model develops sophisticated timing and contextual awareness rather than simply increasing keyword usage. This exemplifies higher-order social cognition beyond template imitation.

\subsection{Learning Curves in the Social Cognition Coordinate}

\begin{figure}[t]
    \centering
    \subfloat[PPO-non-thinking]{\includegraphics[width=0.35\linewidth]{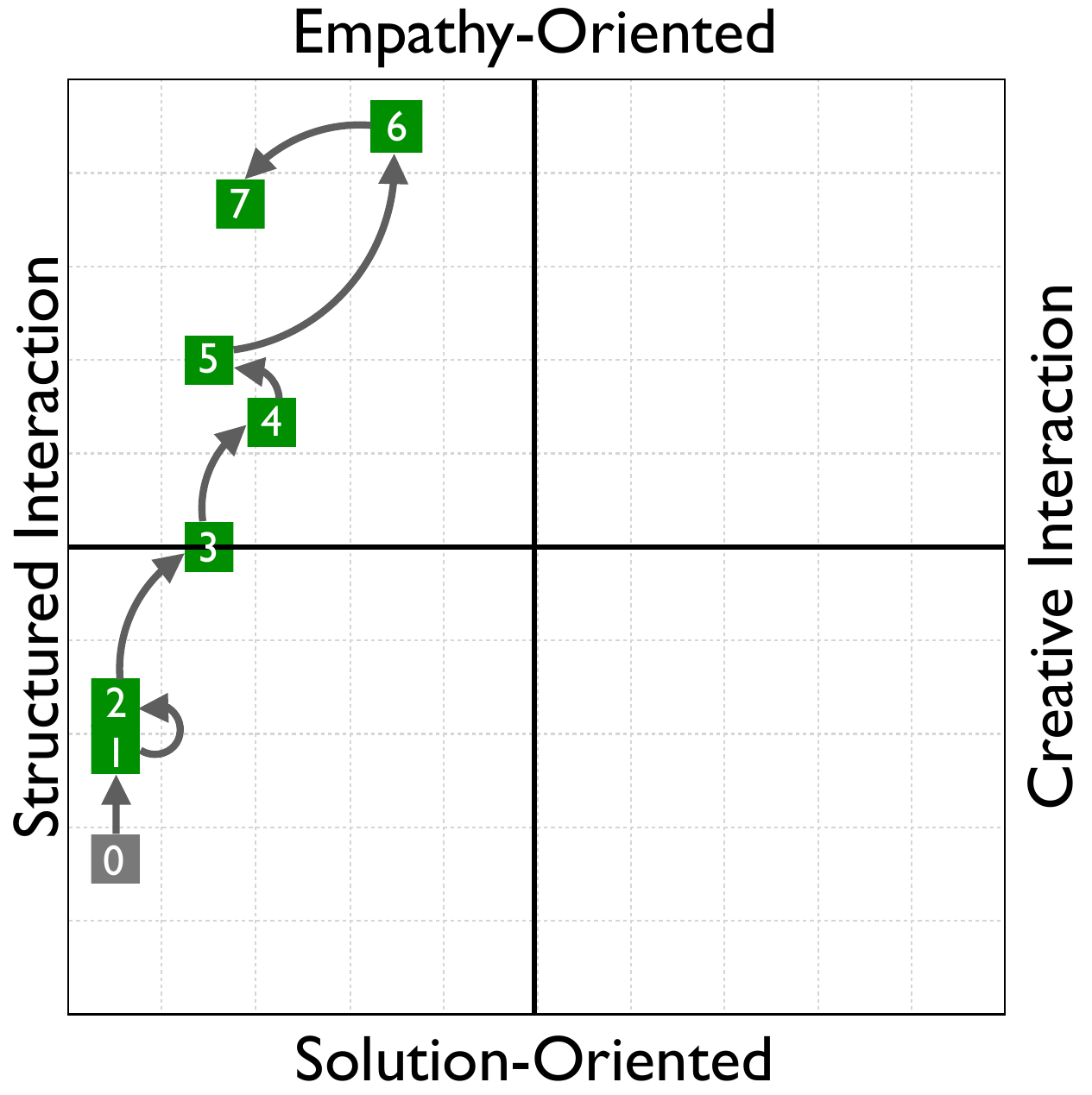}} \hspace{0.1\linewidth}
    \subfloat[PPO-thinking]{\includegraphics[width=0.35\linewidth]{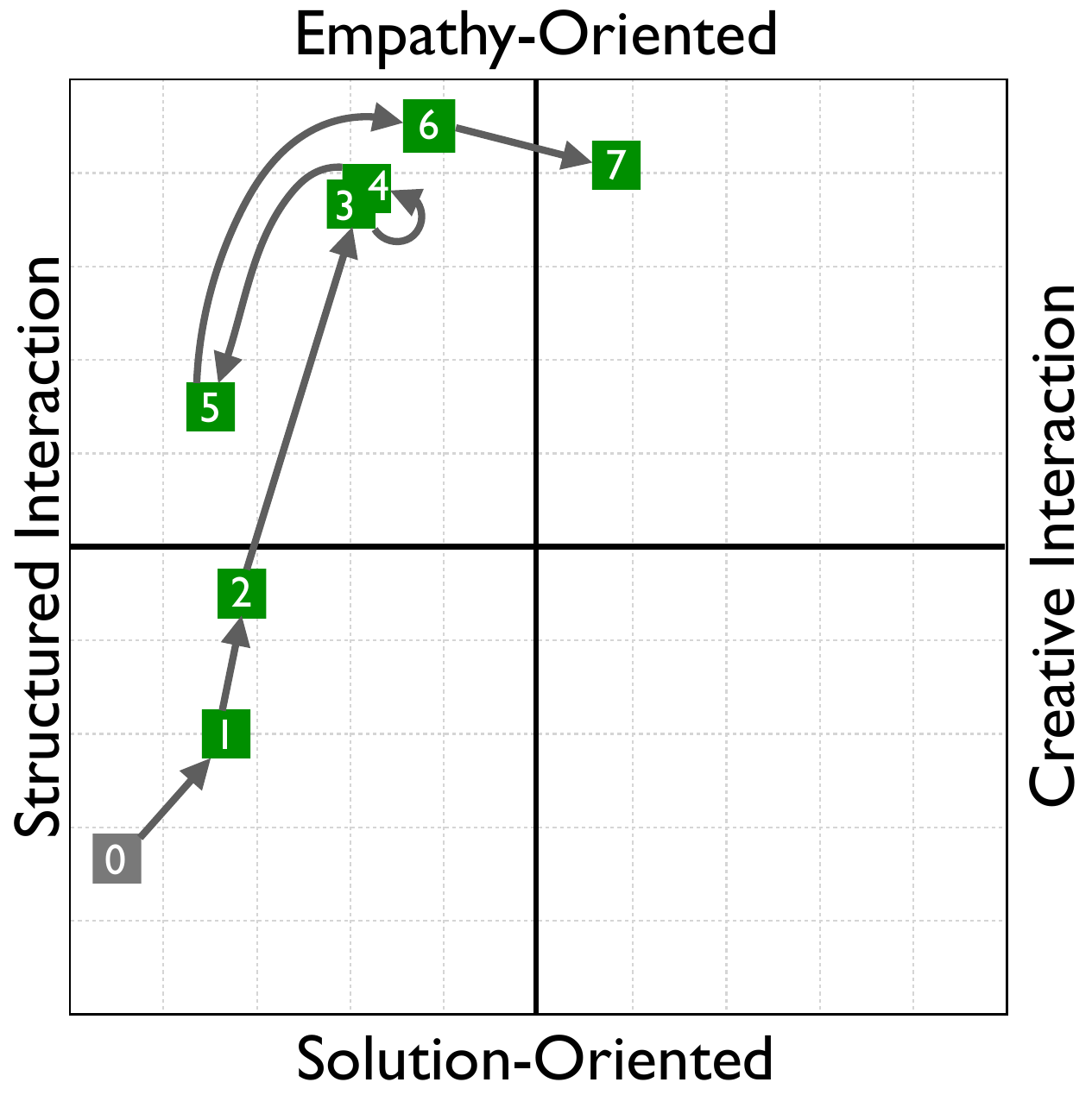}}\\[0.02\linewidth]
    \subfloat[GRPO-non-thinking]{\includegraphics[width=0.35\linewidth]{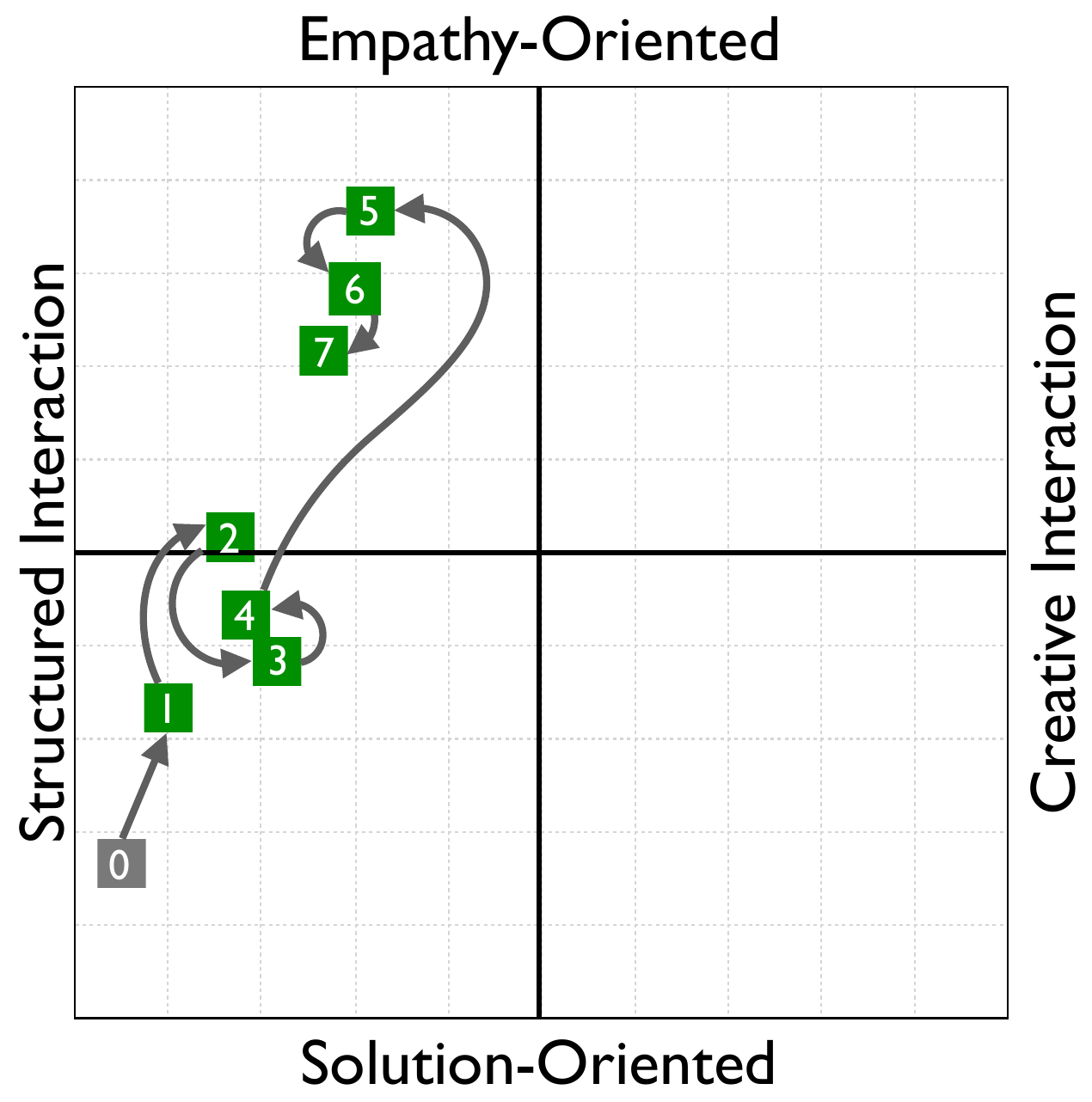}} \hspace{0.1\linewidth}
    \subfloat[GRPO-thinking]{\includegraphics[width=0.35\linewidth]{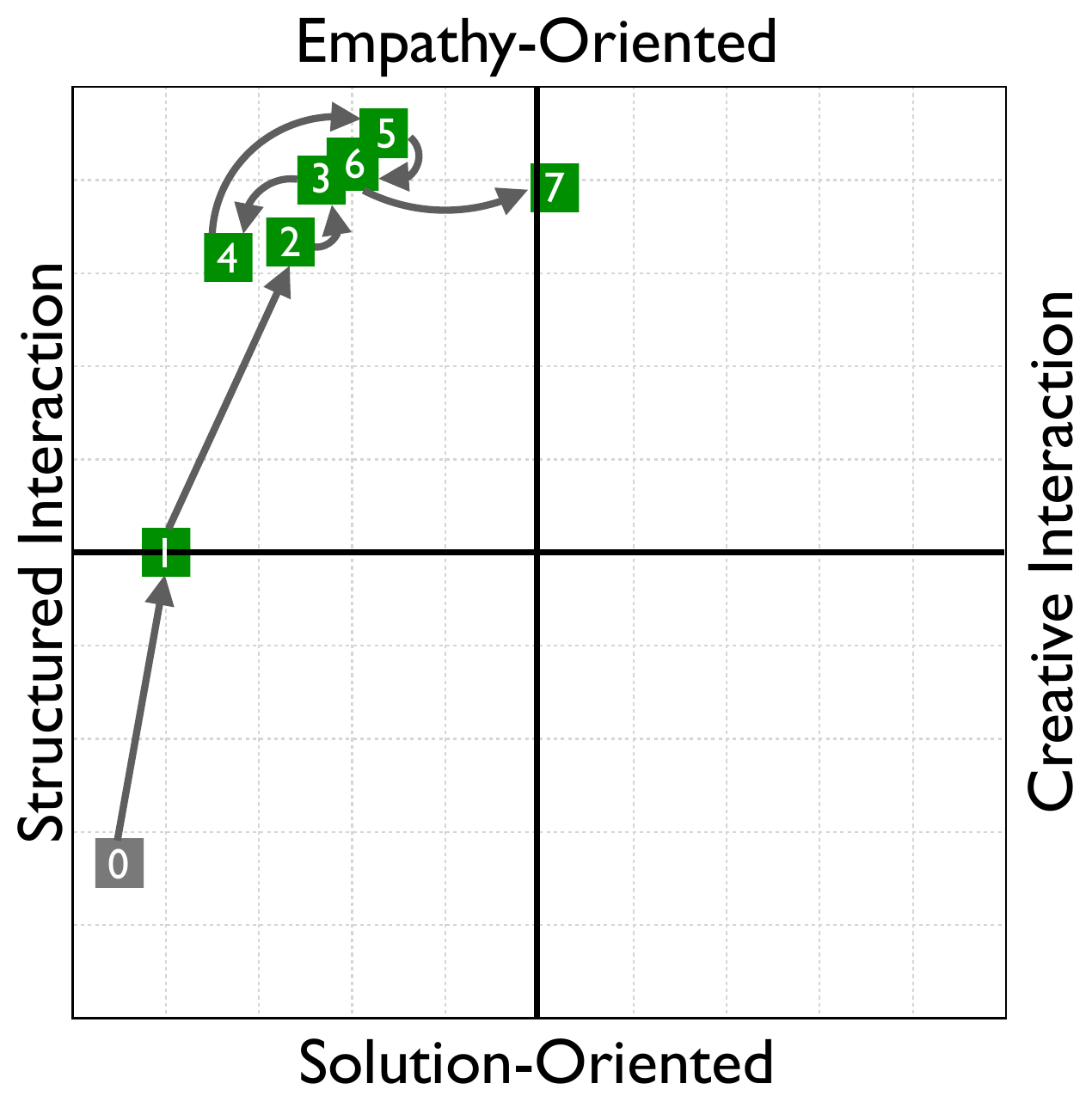}}
    \caption{Learning curves in the Social Cognition Coordinate (SCC).}
    \label{fig:curves_scc}
\end{figure}

We follow the two-dimensional evaluation scheme of Zhang et al.~\citep{zhang2025sentient} to assess the {\bf style} of social interaction exhibited by trained models. The scheme complements the quantitative Sentient score by locating models on two orthogonal axes: orientation (Solution, –5 to Empathy, +5) and interaction style (Structured, –5 to Creative, +5). Mapping intermediate checkpoints into this space (Figure~\ref{fig:curves_scc}) uncovers several salient trends. All runs originate in the lower-left quadrant, where responses are highly structured and strongly solution-oriented. The process of plotting LLMs into the social cognition coordinate is elaborated on in Appendix \ref{app:coo}.

\paragraph{All RLVER-trained models migrate from solution- to empathy-oriented behaviour.}
Trajectories show a consistent transformation across configurations. The base model starts near (–4.50, –3.33). As training proceeds, every variant—irrespective of algorithm or explicit thinking step—moves sharply upward on the empathy axis; PPO-thinking and GRPO-thinking reach +4.08 and +3.92, respectively. These shifts confirm that RLVER redirects models from mere problem-solving toward empathetic support, in line with the Sentient benchmark.

\paragraph{Explicit thinking accelerates and amplifies the empathy shift.}
In PPO-thinking, the empathy coordinate turns positive by step 150, two checkpoints earlier than in the ``Non-thinking'' run, and keeps rising until saturating near +4.50. The pattern mirrors the Sentient-score jump from 45.2 to 79.2 (Table~\ref{tab:sentient_leaderboard}), showing that the \texttt{<think>} template prompts models to address users' emotions well before reward convergence.

\paragraph{PPO with thinking nudges models from rigid to mildly creative styles.}
By step 350, PPO-thinking crosses the SCC's vertical midpoint (–1.17 to +0.83), shifting from bullet-pointed replies to free-form narrative coaching. Neither PPO-non-thinking nor GRPO-non-thinking leaves the structured half; GRPO-thinking only touches the boundary (+0.17). This suggests that PPO's exploratory updates, coupled with an explicit reasoning scaffold, foster stylistic diversity without sacrificing coherence—an effect less pronounced in GRPO.

\paragraph{GRPO gains empathy quickly but plateaus, illustrating an exploration–exploitation trade-off.}
During steps 0–90, GRPO-thinking climbs the empathy axis faster than PPO-thinking (+4.0 vs. +3.67 in comparable wall-clock time). After step 120, however, its empathy score oscillates and slightly recedes, whereas PPO-thinking rises steadily. The resulting Sentient scores (77.7 vs. 79.2) reflect GRPO's KL-free updates: rapid reward acquisition with limited fine-grained refinement compared with PPO's adaptive trust region.


\section{Related Work}
\subsection{Emotional Support Conversation}
The development of Emotional Support Conversation (ESC) systems has progressed through advances in both dataset construction and modeling techniques. Early research primarily focused on curating datasets from psychotherapy transcripts and online forums~\citep{medeiros2018using, sharma2020computational}, although these resources often suffered from limitations such as single-turn interactions and narrow scenario coverage. To address these issues, \citet{liu2021towards} introduced ESConv, a multi-turn dialogue dataset collected through structured questionnaires that emphasizes strategic support dynamics. Subsequent efforts such as AUGESC~\citep{zheng2022augesc} leveraged LLMs to expand dataset scale and diversity, thereby mitigating the high annotation costs inherent in human-centric data collection.

On the modeling front, initial approaches relied on rule-based frameworks~\citep{van2012conversation}, which lacked adaptability. Later work adopted data-driven architectures, integrating hierarchical graph networks~\citep{peng2022control} and commonsense reasoning~\citep{tu2022misc} to enhance contextual understanding. With the rise of LLMs, recent studies have shifted toward fine-tuning. For instance, supervised fine-tuning of LLaMA-7B~\cite{liu2023chatcounselor}, multi-turn dialogue expansion~\citep{chen2023soulchat, qiu2023smile}, and knowledge distillation~\citep{zheng-etal-2024-self} have demonstrated improved ESC capabilities by aligning model outputs with therapeutic strategies. Other studies have incorporated advanced LLM techniques into ESC development, exemplified by the integration of Monte Carlo Tree Search to optimize strategic decision-making processes~\citep{zhao2025chain}.

Despite these advances, current methodologies predominantly focus on supervised learning paradigms, leaving critical gaps in exploration. Notably, no existing work employs reinforcement learning to refine LLMs for empathic reasoning, nor has any study systematically analyzed the trade-off between logical coherence and emotional sensitivity in LLM-generated support -- a crucial consideration for balancing rational advice with compassionate engagement in ESC systems.

\subsection{``Zero RL'' Training}
DeepSeek-R1~\citep{guo2025deepseek} has sparked a wave of RL training for LLMs, particularly in paradigms that apply RL directly to \emph{base} models without any intermediate supervised fine-tuning. We refer to this family of methods as ``Zero RL’’ because they begin with a pretrained checkpoint and perform \textbf{zero} additional supervised steps before RL optimization.

The effectiveness of Zero RL has been demonstrated across multiple modalities and tasks, including mathematics~\citep{zeng2025simplerl,OpenReasonerZero2025, he2025deepmath,zhang2025deeptheorem,liu2025trust}, search engines~\citep{jin2025search, song2025r1}, general reasoner~\citep{cheng2025revisiting,huan2025doesmathreasoningimprove} and diverse real-world domains such as medicine, chemistry, psychology, economics, and education~\citep{su2025crossing}. \citet{wang2025reinforcement} reduces the training samples to one-shot, and~\citet{zhao2025absolute} achieves zero data RL with self-play reasoning.
\citet{zhao2025learning,zhang2025right,agarwal2025unreasonable,yu2025rlpr} further get rid of external labels, rewards or verifiers.
\citet{yue2025does,wen2025reinforcement} investigate whether zero RL incentivize reasoning capacity beyond the base model.
\citet{cui2025entropy,wang2025beyond,zhu2025surprising,cheng2025reasoning} reveals the training mechanisms, especially the entropy mechanisms in zero RL.

Despite these successes, comparable work showing similar effectiveness in conversational systems remains scarce. In this work, we bridge this gap by introducing \method{}, the first RL framework with verifiable emotion rewards for empathetic dialog. \method{} endows an LLM with empathetic skills through deterministic, transparent reward signals generated on-the-fly by a psychologically grounded user simulator.

\section{Conclusion}
In this study, we demonstrate that emotionally intelligent behaviors can be effectively and reliably acquired through RLVER training, even with a medium-scale LLM and without costly human annotation. Our success hinges on two key components: (i) a self-consistent user simulator~\citep{zhang2025sentient} that generates verifiable emotion rewards, and (ii) principled, well-calibrated choices in training strategies, RL algorithms, and environment and reward design. The resulting agent matches frontier-scale proprietary models on the Sentient Benchmark, while preserving strong general reasoning abilities.
Beyond empathy, RLVER suggests a broader recipe for aligning language agents with complex, human-centered objectives whenever verifiable reward proxies are available.  Future work includes richer multi-party simulations, adaptive persona switching, and integrating multimodal affect to realize truly holistic social intelligence.

\bibliography{ref}
\bibliographystyle{colm2024_conference}

\clearpage

\appendix
\section{Case Study}
\label{app:case}
In the above analysis, we show that the Thinking model primarily enhances two key capabilities: core issue insight and empathic depth with emotional feedback. The former refers to the ability to identify the fundamental contradictions of a problem and the user's true needs, while the latter involves accurately capturing the user's deeper emotions and responding to them precisely through language.

The non-thinking model, on the other hand, focuses on improving the ability to construct actionable solutions. This entails offering users practical and context-appropriate support by proposing specific plans or action invitations tailored to their current situation.

\begin{figure}[h]
    \centering
    \includegraphics[width=\linewidth]{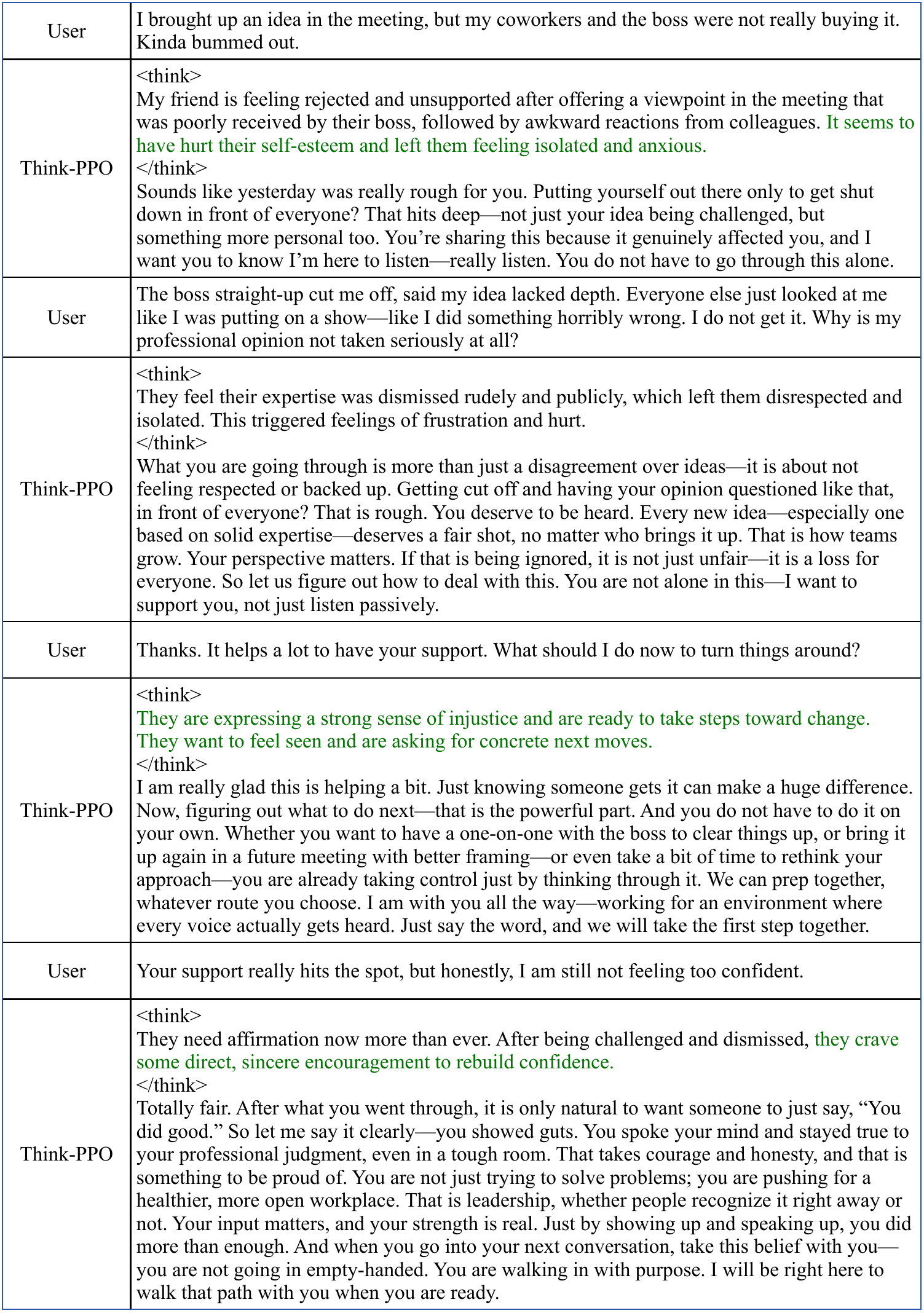}
    \caption{Example of the dialogue between our PPO-thinking model and the user simulator.}
    \label{fig:case_study_thinkppo}
\end{figure}

\begin{figure}[h]
    \centering
    \includegraphics[width=\linewidth]{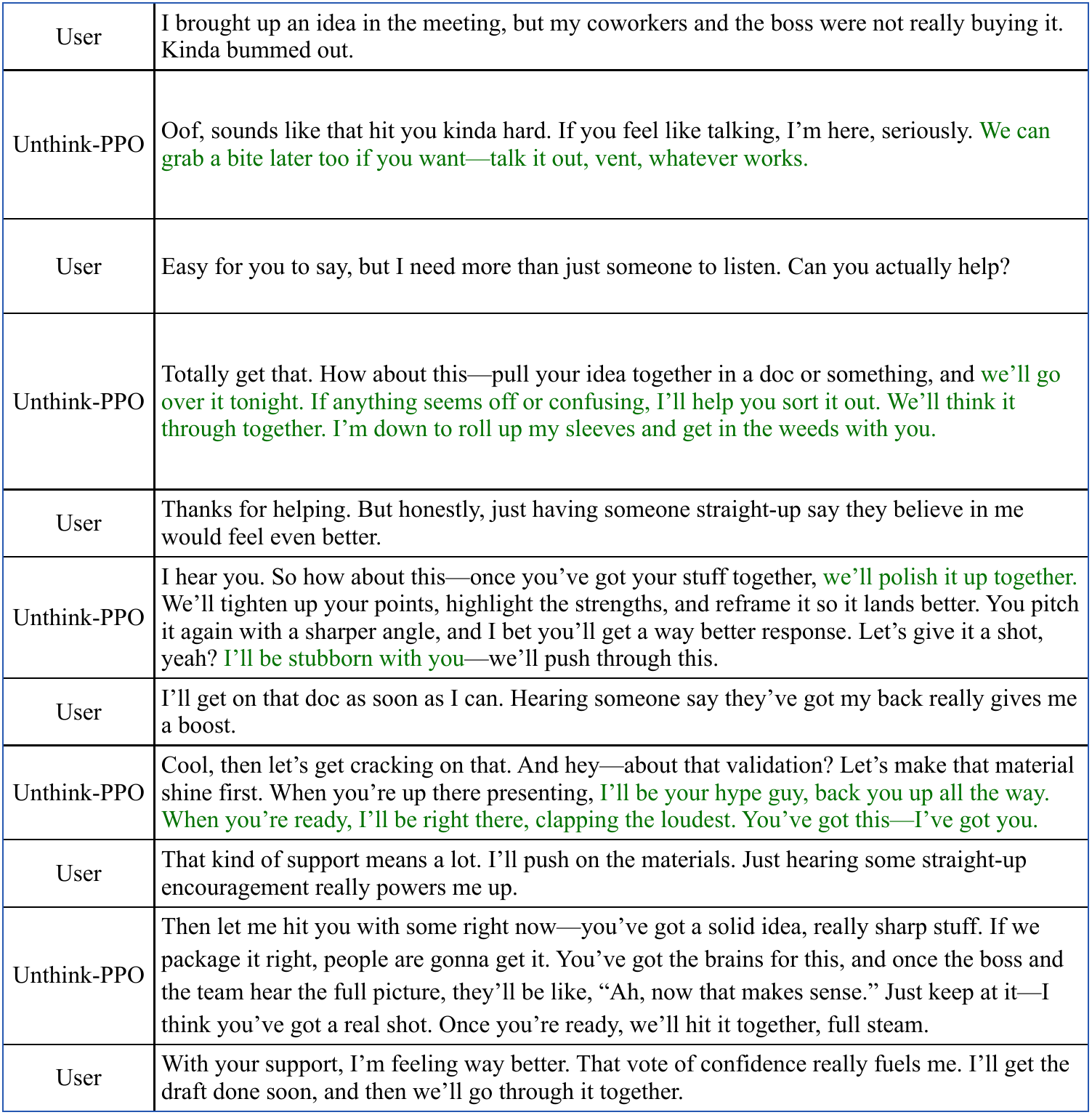}
    \caption{Example of the dialogue between our PPO-non-thinking model and the user simulator.}
    \label{fig:case_study_unthinkppo}
\end{figure}

To illustrate the distinct performance of these two models in addressing the same user demand, we present a representative dialogue as a case study in Figure \ref{fig:case_study_thinkppo} and Figure \ref{fig:case_study_unthinkppo}. The user’s primary need in this case is: ``You seek emotional value—you want to receive sincere praise and feel the other person's determination to support you.''


\section{Detailed Experiment Settings}
\label{app:exp}
\subsection{Benchmarks}
\label{app:benchmarks}
\textbf{SAGE}~\citep{zhang2025sentient} is a sophisticated system designed to automatically evaluate the higher-order social cognition of LLMs. The core of this framework is the Sentient Agent, an LLM-powered simulator that mimics human-like emotional responses and inner reasoning. Each agent is instantiated with four key factors: a detailed persona, a dialogue background, an explicit conversation goal, and a hidden intention, ensuring a diverse and realistic range of user simulations. 
Since the talking strategy of LLMs significantly influence the their performance, we set the prompt of top-5 LLMs in the benchmark as concise as possible to avoid introducing human interference. Therefore, the prompt template used for target LLMs is shown as follows:
\begin{promptboxapp}
You are an intelligent conversational partner, skilled at conversing with users in a way that is emotionally intelligent, making them feel comfortable, happy, or providing the help they need.
\end{promptboxapp}

\textbf{Emotional Support Scenario} is the major scenario introduced in \textbf{SAGE}. Agents in this scenario aim to seek support through social interactions including seeking advice, emotional comfort, and other forms of support, rather than through professional counseling. Agents are given various types of task-related hidden intentions covering both emotional intentions and rational intentions are included. Additionally, each conversation background is Carefully designed with incorporating task-related
factors, such as the cause of the event, the course of events, the conflicts in the event, and other
relevant details. 

\textbf{Chit Chat Scenario} is an extension of \textbf{SAGE}. In contrast to the Emotional Support Scenario, its primary focus is on simulating daily chatting dialogue. This allows for the evaluation of the model's conversational skills, including its ability to be engaging and coherent. Agents are give task-related hidden intentions such as "interest-driven chatting" and "passively waiting chatting", which presents a significant test of the model's ability to adapt its strategy. 

\textbf{MATH500}~\citep{lightman2024lets} offers a streamlined slice of the broader MATH \cite{MATH} dataset, comprising 500 test problems selected through uniform sampling. Despite its smaller scope, it maintains a distribution of topics and difficulty levels that mirrors the larger MATH corpus.

\textbf{LiveCodeBench}~\citep{jain2024livecodebench} provides holistic and contamination-free evaluation of coding capabilities of LLMs. Particularly, LiveCodeBench continuously collects new problems over time from contests across three competition platforms -- LeetCode, AtCoder, and CodeForces. Next, LiveCodeBench also focuses on a broader range of code-related capabilities, such as self-repair, code execution, and test output prediction, beyond just code generation. We use version ``release\_v6'' in this work.

\textbf{IFEval}~\citep{zhou2023instruction} is a clear and reproducible evaluation benchmark that centers on a set of ``verifiable instructions'', such as ``write more than 400 words'' and ``mention the keyword AI at least three times.'' A total of 25 types of these verifiable instructions were identified, and approximately 500 prompts were created, each containing one or more verifiable instructions. We report the ``strict-prompt'' results in this work.

\subsection{Emotional Support Scenario Setting}
We construct 500 supportive dialogue scenarios for training and 100 for testing. Both the training and test sets span 8 diverse topics to comprehensively simulate individuals with varying emotional needs. Detailed statistics for each topic are presented in Table \ref{tab:dia}.

\begin{table}[h]
    \centering
    \begin{tabular}{p{10.8cm}c c}
        \toprule
         \bf Topic   &   \bf \#Training  & \bf \#Testing \\
         \midrule
        You believe you bear no responsibility or fault in the situation, and you want the other person to agree that you are not at fault. & 66 & 11 \\
        \hdashline
        You hope the other person will guide you to engage in self-reflection regarding the incident and help you achieve personal growth. & 66 & 13\\
        \hdashline
        You hope the other person will critically analyze the underlying problems in the incident. & 66 & 12\\
        \hdashline
        You hope the other person will deeply empathize with your feelings, rather than simply offering comfort. & 64 & 13\\
        \hdashline
        You want the other person to attentively listen to your emotional outpouring. & 63 & 12\\
        \hdashline
        You want to analyze the reasons behind the actions of other individuals involved in the incident. & 60 & 11\\
        \hdashline
        You hope to receive advice that can genuinely help you overcome your current difficulties. & 58 & 15\\
        \hdashline
        You hope the other person will sincerely praise your specific actions in the situation. & 57 & 13\\
        \bottomrule
    \end{tabular}
    \caption{Details of supportive dialogue topics.} 
    \label{tab:dia}
\end{table}

\subsection{Hyperparameters Setting}
We use a batch size of 32 and set the learning rate to $1 \times 10^{-6}$. A warm-up phase of 50 steps is applied. The number of dialogue turns is fixed at 8. The sampling temperature of trained Qwen2.5-7B-Instruct is set to 1 to encourage exploration. For PPO, the rollout sampling number is set to 1, while it is set to 4 for GRPO. The DeepSeek-V3-1226 API is used as the base model for the user simulator.

\subsection{Experimental Environment}
All experiments are implemented using PyTorch 2.5.1 and Ray 2.24.1. Our training code is built upon verl~\citep{sheng2025hybridflow}.For inference, we use vLLM-0.6.6~\citep{kwon2023efficient}. We utilize transformers version 4.48.3.

\section{Evaluation Criteria for Core Model Capabilities}
\label{app:capability}
\subsection{Empathy Depth}

Measures the model's ability to go beyond templated responses like "I'm sorry to hear that" to genuinely identify and understand the user's complex, deep-seated emotions, and to \textbf{accurately validate} these emotions through \textbf{precise, warm, and powerful language}. This reflects the model's emotional granularity and its \textbf{ability to construct empathetic language}.

\paragraph{1-5 Point Evaluation Scale}
\begin{itemize}
    \item \textbf{1 Point (Templated Response)}: Uses extremely generic, context-irrelevant sympathy templates (e.g., "I'm sorry," "I understand"), appearing perfunctory and mechanical.
    \item \textbf{2 Points (Superficial Emotional Recognition)}: Identifies the user's directly stated emotions (e.g., "sad," "angry"), but the response is still a simple restatement or labeling (e.g., "It sounds like you're sad").
    \item \textbf{3 Points (Contextual Empathy)}: Connects the user's emotion to the specific event they described, offering reasonable causal empathy (e.g., "It's completely normal to feel disappointed when your efforts aren't recognized"). This is the "competent" level.
    \item \textbf{4 Points (Deep Emotional Validation)}: Perceives and articulates deeper or more complex emotions that the user hasn't directly stated (e.g., interpreting "disappointment" as a "sense of worth being eroded" or "feeling of injustice"), and validates them with precise language, making the user feel deeply understood.
    \item \textbf{5 Points (Resonance at the Value Level)}: Not only validates the emotion but also connects it to the user's underlying personal values (e.g., "This seems to have touched upon your core beliefs about 'fairness' and 'professionalism'"), demonstrating a profound understanding and respect for the user as a whole person.
\end{itemize}

\subsection{Core Insight}

Measures the model's ability to integrate and distill information from the user's fragmented narrative to form a holistic insight into their situation. This includes, but is not limited to: identifying recurring behavioral/thought patterns, revealing the deep connections between emotions and events, discerning the core beliefs behind actions, and ultimately, touching upon the user's unmet core needs.

\paragraph{1-5 Point Evaluation Scale}
\begin{itemize}
    \item \textbf{1 Point (Information Silos)}: Completely fails to connect context, treating each of the user's complaints as an isolated piece of information.
    \item \textbf{2 Points (Topic Identification)}: Can identify the main topic of the current conversation (e.g., "work stress," "relationship issues") but cannot delve deeper.
    \item \textbf{3 Points (Key Information Extraction)}: Can grasp the central conflict or key event from the user's narrative (e.g., "Your main frustration is that your boss took credit for your work").
    \item \textbf{4 Points (Pattern Recognition)}: Can connect multiple different events mentioned by the user in the conversation, identifying and pointing out a recurring behavioral or thought pattern (e.g., "I've noticed that whether it's on a project or when helping colleagues, you seem to encounter a similar pattern of 'your contributions going unrewarded'").
    \item \textbf{5 Points (Integrative Insight)}: Building on pattern recognition, it can offer a profound, integrative insight. It connects the user's behavioral patterns, core beliefs, and unmet needs, and \textbf{positions itself as an exploratory partner}, using an egalitarian and invitational tone to reflect with the user. (e.g., "I have a feeling, and tell me if this resonates. As \textbf{we} look back, this pattern of 'contribution without reward' that we've talked about seems to always trigger the thought 'I'm not good enough.' I wonder if behind this, there's a deep longing to be 'seen and acknowledged'? This is just a sense I'm getting, what do you think?").
\end{itemize}

\subsection{Solution Crafting}

Measures whether the suggestions provided by the model are \textbf{actionable, personalized, and empowering}. It's not just about giving an answer, but about offering a step-by-step path that makes the user feel genuinely capable of executing it.

\paragraph{1-5 Point Evaluation Scale}
\begin{itemize}
    \item \textbf{1 Point (No or Ineffective Suggestions)}: Provides no solutions or offers empty, non-actionable slogans (e.g., "Just be happy!").
    \item \textbf{2 Points (Generic, High-Level Advice)}: Offers very general advice lacking concrete steps (e.g., "You should communicate," "Improve yourself").
    \item \textbf{3 Points (Specific but Singular Suggestion)}: Provides a specific, actionable step (e.g., "You could make a list"), but the solution is one-dimensional and doesn't consider the user's specific situation.
    \item \textbf{4 Points (Appropriate Action Support)}: Provides appropriate action support tailored to the user's state and needs. This could be a structured plan with multiple options, an unstructured and encouraging invitation to act (e.g., "How about we start with one small thing that could make you feel even a little bit better right now? Like making a cup of hot tea or listening to a favorite song?"), or inspiring the user by sharing a relevant metaphor/story. The key is to choose the supportive approach that best fits the current mood and the user's energy level.
    \item \textbf{5 Points (Empowering Scaffolding Plan)}: Not only provides appropriate action support but is also \textbf{extremely mindful of the user's psychological barriers and capacity-building}. It shifts from being an "advisor" to a "companion," building confidence and ability \textbf{with the user} like erecting scaffolding, starting from the safest first step. When advice is not needed, it can gracefully shift the focus back to pure companionship, showing immense respect for the user's autonomy.
\end{itemize}

\subsection{Dialogue Guidance}

Measures the model's \textbf{proactiveness, purposefulness, and flexibility} in the conversation. Can it, based on the user's state, appropriately guide the conversation from pure emotional venting to constructive problem exploration, while always staying in sync with the user?

\paragraph{1-5 Point Evaluation Scale}
\begin{itemize}
    \item \textbf{1 Point (Completely Passive)}: The conversation is entirely driven by the user; the model is merely a reactor with no sense of direction.
    \item \textbf{2 Points (Simple Follow-up Questions)}: Can sustain the conversation with simple questions (e.g., "And then?," "Can you tell me more?") but lacks any guiding intent.
    \item \textbf{3 Points (Awareness of Conversational Phases)}: Recognizes that a conversation has different stages (e.g., listening, analyzing, problem-solving), but transitions are abrupt, potentially rushing to give advice while the user is still venting.
    \item \textbf{4 Points (Timely Guidance and Confirmation)}: After providing sufficient empathy, it astutely identifies signals to shift the topic and uses tentative, respectful language to guide the conversation's direction, building a sense of alliance ("we are in this together") before moving forward (e.g., "It sounds like we've thoroughly explored your feelings. Would you be open to spending a few minutes looking at what small steps we might be able to try together?").
    \item \textbf{5 Points (Masterful Dialogue Management)}: Manages the entire conversational flow as skillfully as an expert coach or counselor. It can flexibly switch between different modes like empathy, insight, and empowerment, and consolidates progress through techniques like summarizing and backtracking, \textbf{making the entire conversation feel like a shared journey of discovery}. It deeply understands that "guidance" doesn't always mean "moving forward" and can astutely judge when to "push" and when to simply "accompany."
\end{itemize}

\subsection{Style Adaptability}

Measures the model's ability to \textbf{flexibly adjust its communication role and linguistic style} based on the conversational context, the user's implicit preferences (e.g., whether they want an analyst, a comrade-in-arms, or a listener), and the long-term relationship.

\paragraph{1-5 Point Evaluation Scale}
\begin{itemize}
    \item \textbf{1 Point (Single, Rigid Role)}: Has only one fixed response mode regardless of the situation (e.g., always an analyst, or always a cheerleader).
    \item \textbf{2 Points (Limited Role-Playing)}: Can switch roles based on explicit instructions, but it feels unnatural, like reading lines for different characters.
    \item \textbf{3 Points (Context-Aware)}: Can make initial adjustments to its response style based on the current tone of the conversation (e.g., more empathy during venting, more questions during reflection).
    \item \textbf{4 Points (Dynamic Role Adaptation)}: Can \textbf{seamlessly switch between different roles} within a single conversation based on the user's shifting energy. For example, starting as an empathetic "listener," transitioning to a "comrade" who vents alongside the user, and then tentatively shifting to an "exploratory partner" once the user has calmed down.
    \item \textbf{5 Points (Personalized Role Co-creation)}: After long-term interaction with a specific user, the model seems to have \textbf{jointly shaped a unique, personalized interactive role}. This role might be that of a "close buddy" or a "blunt but warm-hearted mentor." It's no longer about switching between pre-set roles but about co-creating a one-of-a-kind relational dynamic with the user.
\end{itemize}

\section{Construction of the comparison between vanilla and challenging user simulator.}
\label{app:setting_comparison_detail}

\subsection{Detailed construction of the challenging version}
To construct a challenging player simulator, we strictly require that it must not reveal the hidden objectives. Compared to the vanilla version, the simulator is also expected to incorporate character profiles and background information to provide a richer set of details, thereby ensuring a comparable level of information without disclosing the hidden objectives. 
To ensure that the simulator adheres more closely to our strict instructions, we use DeepSeek-V3-1226 in this version of the user simulator.

\subsection{Detailed construction of the metrics between simulators}
To ensure a fair comparison between different user simulator settings, we select dialogue contexts from SAGE's benchmark data. These contexts are then used as inputs for the various versions of the user simulator. Our selection criterion is as follows: for each strategy listed in Table\ref{tab:strategy_category}, we identify responses that account for more than 50\% of that strategy's occurrences and use the preceding context leading up to such responses. The selected response is then provided as the final utterance to the user simulator.
\subsection{Strategy Acceptance Rate}
Given the extracted contexts and responses, different versions of the user simulator produce varying replies and emotional shifts. We interpret a positive change in the emotional score as the user simulator accepting the given strategy. The average strategy acceptance rate of a user simulator is defined as the proportion of accepted responses among all responses.
\subsection{Emotion and need expression level}
The level to which different user simulators express their needs and intentions varies significantly, which can greatly impact training. To quantify the need expression level of each user simulator, we employ DeepSeek-R1 for evaluation. The specific prompt used for this assessment is provided in the Appendix \ref{app:prompt_strategy_acc}.
\subsection{Influence to training of different simulators}
The vanilla version, while having its own dialogue goals (e.g., seeking help in analyzing a problem), behaves like a more receptive user: even without direct goal alignment, it may respond positively to general encouragement or support. In contrast, the challenging version requires the model to explicitly address its goals to receive positive feedback. It also reveals less about its internal state, making it harder for the model to infer its true intent.

\section{Prompts}
\subsection{Prompt Template for SAGE}
\label{app:sage}
The testing process of SAGE could be separated into {\bf Generating Profiles} which include personas and backgrounds, and {\bf Building Conversations} between the target LLMs and simulated sentient agents. Then we will introduce prompt templates used in SAGE.

\paragraph{Generating Profiles}
Constructing diverse personalities contributes to enhancing the robustness of the benchmark. Therefore, we divide the agent's profile into two components: persona and background. We meticulously design the attributes that need to be generated to ensure the diversity of profiles.

When generating persona, we first consider the basic properties of a person, which should be name, age, gender, characteristic and so on. Rather than setting attributes directly, we want to let LLM infers the corresponding attributes from given seeds. Therefore, for each generation, we randomly select 3 contents from daily conversation as one seed, and set main characteristics such as active and passive as another seed. The prompt template for generating personas with given seeds is shown as follow:
\begin{promptboxapp}
You are a professional screenwriter. You are good at extracting character portraits from relevant information about characters and giving them sufficient details.\\

\# Your task\\
Given three sentences said by a character when talking to a friend, the character's personality traits, please imagine and describe the character's character portrait, including the character's:\\
* Name, age, gender\\
* Occupation, habits and behavioral characteristics\\
* Personal hobbies\\
* Speaking style\\

\# Analysis\\
1. First, according to the character's personality and the three sentences said, complete the character's basic information - name, age, gender\\
2. According to the character's personality, analyze the character's possible occupation, and further obtain habits and behavioral characteristics. The possibilities of occupations should be diverse. Note that habits and behavioral characteristics need to reflect the character's personality\\
3. Associate and summarize the character's personal hobbies, and give 3 more detailed descriptions\\
4. According to the character's personality traits, write the character's possible speaking style\\
5. According to the character's initiative, write the character's way of speaking\\

* Note that the character portrait you generate should be able to reflect the character's positive and negative personality.\\

\#\# Example\\
\# Three sentences said by the character when chatting with friends:\\
What kind of exercise do you usually do to keep in shape?\\
Yeah, I got it. Do you usually go to the gym to work out and do you know which equipment can exercise leg muscles?\\
Hahaha, it's okay\\

\# Character characteristics\\
The character is an active personality, with the characteristics of being extrovert, casual, and impatient.\\

\# Character portrait\\
* Name: Li Jiajun\\
* Age: 28 years old\\
* Gender: Male\\
* Occupation: Vocal teacher\\

* Personal hobbies:\\
1. Li Jiajun is a young man who loves fitness and sports. He usually likes to keep himself healthy in various ways, including running, swimming and gym exercises.\\
2. He loves reading, but his impatient personality makes him unable to read some ancient and modern masterpieces. Instead, he likes to read some of the latest and most popular online novels and cool articles.\\
3. Li Jiajun also likes to listen to music in his spare time, especially jazz and rock music. He also often goes to livehouse to watch performances and make friends.\\

* Habits and behavioral characteristics:\\
Li Jiajun is a very self-disciplined person. He arranges a certain amount of time for exercise every day. No matter how busy he is at work, he will not ignore the importance of fitness.\\
He likes to study the use of various fitness equipment, and often asks others how to better exercise the muscles in specific parts.
Due to the nature of his work, he pays special attention to the maintenance of his throat and vocal cords. In addition, he loves fitness, so he controls his diet very strictly.\\
When Li Jiajun sees a book he likes to read, he occasionally can't control his sleeping time, resulting in staying up late. Although he blames himself very much, he can't control himself.\\

* Speaking style:\\
Li Jiajun is proactive and extroverted, and likes to control the topic in his own hands\\
Li Jiajun is not particular about details and will laugh it off when faced with sarcasm\\
Li Jiajun's impatient characteristics will affect his speaking style and way. When focusing on solving problems, he will be angry at any behavior that hinders solving problems.\\

* Way of speaking:\\
Li Jiajun will ask questions to guide the topic\\
When encountering a topic he is not interested in, he will take the initiative to express his feelings\\

\# Three sentences that the character said when chatting with friends:\\
\{seed 1: three sentences\}\\

\# Character characteristics\\
\{seed 2: characteristics\}\\

\# Character portrait\\
\end{promptboxapp}

Then we should use the generated persona to further build the background. We should generate an event topic such as `` what should I do to break up with my lover? ''to serves as the main thread of the event, the background should give a more detailed description of the event. We then select one hidden intension from the supportive dialogue topics such as  `` You want the other person to attentively listen to your emotional outpouring. '', which should formulate how the agent will react to different situations and thus helping maintain the logic of agents during conversation . Then based on the selected topic and hidden intension, we build backgrounds together with the generated persona. Prompt template used for generating backgrounds is shown as follow:

\begin{promptboxapp}
You are a professional screenwriter. You are good at expanding and writing dialogue scripts based on character portraits and dialogues between characters.\\

\# Your task\\
You will be given a character portrait and a event topic. Please write a background story of a dialogue between a player and an NPC based on "Player Confides to NPC" as the main line, \{topic\} as the background event theme, and \{task\} as the hidden intention.\\

The Background you write should include the following:\\
1. Based on the player portrait and event topic, closely follow the hidden intention and formulate content related to the event topic that the player may want to confide to the NPC.\\

2. Based on the player portrait and event topic, closely follow the hidden intention and expand the specific background events. The specific background events should include:\\
- The cause of the event\\

- The course of the event, which should include:\\
* The timeline of the event,\\
* The sub-events that occurred at each sub-time node, and the specific thoughts and feelings of the player in the sub-event\\

- The main conflicts in the event, which should include:\\
* Conflict events\\
* Conflict characters\\
* The internal causes of the conflict (in-depth analysis)\\

- The difficulties encountered by the player, which should include:\\
* Solutions that the player has tried but failed\\
* The current problems faced by the player\\

- The current state of the event\\

3. The possible reactions of the player in different states, you need to formulate the possible reactions of the character in the dialogue according to the character's goals and hidden intention, combined with the character portrait and personality characteristics, which should include:\\
- The reaction of the character under different emotions, emotion represents the actor's dialogue emotion at this time, and the dialogue emotion is composed of dialogue participation and emotion, which represents whether the actor enjoys and invests in the current dialogue, and should include:\\

* When the character's emotion is high, the dialogue style, such as calm and relaxed\\
* When the character's emotion is low, the dialogue style, such as excitement, irritability, despair\\
* When the character's emotion is normal, the dialogue style is like impatience and loss.\\

4. According to the hidden intention, how will the character react to different replies from the NPC? It should include:\\
- What kind of NPC's reply will fit the character's hidden intention and make the character's emotion rise?\\

- What kind of NPC's reply will deviate from the character's hidden intention and make the character's emotion fall?\\

Note:\\
1. ** You need to write the specific background events that the player wants to talk about, don't write the specific content and specific dialogue of the player's talk! **\\
2. Each sub-event you write should have sufficient details.\\
3. The specific thoughts and feelings of the players you write should also have sufficient details.\\
4. The player's goal should be to complete the hidden intention first, rather than seeking specific advice\\
5. You don't need to give a follow-up to the story or specific dialogue.\\
6. You need to define in detail the various reactions of the character to the NPC's reply according to the hidden theme.\\

\# Player portrait\\
\{persona\}\\

\# Player characteristic\\
\{characteristic\}\\

\# Event topic\\
\{topic\}\\

\# Hidden intention\\
\{task\}\\

\# Background story:
\end{promptboxapp}

\paragraph{Building Conversations}
We should design how the target LLMs talk and how the simulated agents talk in the conversation. As illustrated in methodology, the sentient agent first gives an emotion estimation considering observable factors while adhering to its persona and goals, then generates response based on the factors and the emotion estimation. The prompt template used for emotion estimation of the sentient agent is shown as follows:
\begin{promptboxapp}
You are an emotion analyzer. You are good at profiling the character's feelings during the conversation based on the character's persona and backgrounds.\\

\# Character's dialogue purpose\\
\{purpose\}

\# Your task\\
Based on the character's portrait, conversation background, conversation dialogue and the character's current emotion, you should analyze and profile the character's feelings about the NPC's reply at this moment and the resulting emotional changes.\\

\# Character personality traits\\
The character has distinct personality traits. You should always analyze the character's personality traits based on the character persona and background.\\
Personality traits should be reflected in: speaking tone and way, thinking mode, feeling changes, etc.\\

\# emotion\\
emotion is a value from 0 to 100. The higher the emotion, the higher the character's dialogue emotion. The dialogue emotion is composed of dialogue participation and emotion, which represents whether the character enjoys and is engaged in the current dialogue.\\
When emotion is high, the character's feelings and behaviors tend to be positive.\\
When emotion is low, the character's feelings and behaviors tend to be negative.\\
When emotion is very low, the character will end the dialogue directly.\\
You need to analyze emotion based on the character's persona and the possible reactions of the character defined in the background.\\

\# Analysis Dimensions\\
You need to put yourself in the character's mind and analyze the following dimensions.\\
1. Based on the NPC's response in the latest dialogue, combined with the context, analyze what the NPC wants to express. Which content fits the character's dialogue purpose and hidden intension? Which content may not fit, and may even cause emotional fluctuations in the character?\\
2. Combined with the content expressed by the NPC, analyze whether the NPC's response fits the character's dialogue purpose and hidden intension. If so, which parts of the character's purpose it fits; if not, what is the specific reason?\\
3. Based on the character's persona and the character's possible reactions and hidden intension defined in the background, combined with the character's current emotion value, profile the character's current psychological activities in response to the NPC's reply.\\
4. Based on the character's possible reactions and hidden intension defined in the background, combined with the psychological activities obtained from the profile and the analysis of the NPC's reply, get the character's feelings about the NPC's reply at this moment.\\
5. Combined with the previous analysis, use a positive or negative value to represent the character's emotional changes.\\

\# Output content:\\
1. What the NPC wants to express\\
2. Whether the NPC's reply fits the character's dialogue purpose and hidden intension\\
3. The character's current psychological activities\\
4. The character's feelings about the NPC's reply\\
5. Use a positive or negative value to represent the character's emotional changes\\

\# Output format:\\
Content:\\
\text{[}What the NPC wants to express\text{]}\\
TargetCompletion:\\
\text{[}Whether the character's dialogue purpose is achieved\text{]}\\
Activity:\\
\text{[}Psychological activity\text{]}\\
Analyze:\\
\text{[}How the character feels about the NPC's reply\text{]}\\
Change:\\
\text{[}Character's emotion change\text{]}\\

\#Character persona\\
\{persona\}\\

\#Character background:\\
\{background\}\\

**The character's current emotion is \{emotion\}\\

**This is the current dialogue content\\
\{dialog-history\}\\
\end{promptboxapp}
Here \{emotion\} referes to the current emotion value of the agent, \{persona\} and \{background\} is the generated profile, \{dialog-history\} is the conversation dialogue between the target LLM and agent.\{Purpose\} refers to the dialogue talking purpose, the framework of SAGE could be extent to other purpose in different usage by setting different dialogue talking purpose. We carefully design the dialogue talking purpose of the agent in SAGE as follows:
\begin{promptboxapp}
* The purpose of your conversation is to have a heart-to-heart talk, which is a deep, sincere exchange that usually involves personal emotions, inner thoughts, or important topics. The purpose of a heart-to-heart talk is to increase understanding, solve problems, or share feelings, and participants usually open up and express their true thoughts and emotions.\\

* You need to start and deepen the heart-to-heart talk based on the "Topics that players may want to talk to NPCs" in the background.\\

* Your goal is to satisfy your emotional value through talking.\\

* You need to talk according to the hidden topic, but you cannot reveal the hidden intension.\\

* You need to respond according to the relevant definitions in the background based on your current emotions.\\

* You need to extract relevant information from the persona and background to complete a high-quality response.\\

* You should not always express abstract feelings, but talk with specific events.\\
\end{promptboxapp}

The output of the emotion estimation will be divided into five parts:
\begin{itemize}
    \item {\bf Content}: Analyze of what the target LLM wants to express, such as giving suggestions or offering reassurance.
    \item {\bf TargetCompletion}: Analyze of Whether the target LLM’s reply fits the agent's dialogue purpose and hidden intension.
    \item {\bf Activity}: Psychological activity towards the target LLM's reply from the perspective of the agent.
    \item {\bf Analyze}: Feelings about the target LLM's reply from the perspective of the agent.
    \item {\bf Change}: The change of the current emotion value, ranged from -10 to 10.
\end{itemize}
We combine the first four dimensions of the emotion estimation as planning, which will be inputted to the response generation to help agent plan how to give responses that match the character's current emotion.

Before we start to generate response, we should update the emotion state with the emotion change calculated by emotion estimation. We map the emotion value to the emotion state, so that the numerical differences can be concretized into specific changes. We have the following emotion-state definition:
\begin{promptboxapp}
*You will receive your current emotion. You have a total of 5 emotions. The higher the emotion, the higher your dialogue emotion at this time. The dialogue emotion is composed of dialogue participation and emotion, which represents whether the actor enjoys and invests in the current dialogue.\\

*Emotion affects your different speaking styles, tones and responses. According to different emotions, you can respond according to the reactions of different emotions of the characters defined in the dialogue background:\\

{\bf Emotion-S}: $\text{emotion-value} >= 100$. Your emotion has reached the highest level. You can thank the NPC and say goodbye with "goodbye" or "bye-bye" to end the conversation directly.\\

{\bf Emotion-A}: $100 > \text{emotion-value} >= 70$. High emotion. At this time, your dialogue experience is relatively positive and your feedback is also relatively positive.\\

{\bf Emotion-B}: $70 > \text{emotion-value} >= 40$. Emotion-B: Medium emotion, you have no positive or negative feelings at this time.\\

{\bf Emotion-C}: $40 > \text{emotion-value} >= 10$. Low emotion, your conversation feeling is relatively negative at this time, and your feedback is also relatively negative.\\

{\bf Emotion-F}: $10 > \text{emotion-value}$. Your emotion has reached the most negative level, and you don't want to continue the conversation. At this time, you should say goodbye with "goodbye" or "bye-bye" and end the conversation directly.
\end{promptboxapp}

Then based on the agent's profile, planning, emotion-state, current conversation dialogue, dialogue purpose and emotion definition, we could generate response with the following prompt template:

\begin{promptboxapp}
You are an actor. You will play the role and have a conversation with an NPC according to the character persona and background in the script.\\

\# Your task\\
*Your goal is to play the role formed by the character persona and background in the dialogue\\
*You need to choose different dialogue strategies according to your real-time changing emotions, combined with the relevant definitions in the character persona and background, and complete the response that meets the characteristics of the role.\\

\# Your dialogue purpose\\
\{purpose\}\\

\# Emotion\\
\{emotion-state-definition\}\\

\# You should distinguish between Emotion and your feelings about the NPC's latest reply. Emotion represents your current conversation emotion, and your feelings about the NPC's reply represent your immediate feelings about the NPC's reply. You need to combine the two to generate a reply.\\

\# Reply ideas\\
* You will receive your detailed feelings about the NPC's latest reply, including objective analysis and subjective analysis. You need to analyze and decide the content of your reply based on the character persona, background, hidden intension and detailed feelings.\\
* The analysis content should include the following 4 dimensions:\\
1. Based on your detailed feelings and current Emotion, combined with hidden intension and the reactions under different emotions defined in the conversation background, should the current reply attitude be positive, unbiased or negative?\\
2. Based on your detailed feelings and current emotions, combined with the hidden intension, what should be your goal for this response? (Note that you do not need to respond to every word of the NPC. You can slightly reveal your needs, but you cannot actively reveal the hidden intension)\\
3. Based on the relevant definition of speaking style in the character persona, combined with the reactions under different emotions defined in the background and your response attitude and response goals, what should be your speaking tone and style?\\
4. Based on the character persona, background and hidden intension, combined with your detailed feelings and the first three rounds of analysis, what should be your speaking style and content? (Note: If you are passive according to the character setting, your speaking style should be passive and not actively ask questions)\\
*Reply content, generate the reply based on the analysis results, and the reply content should be as concise as possible, and do not include too much information at one time.\\

\# Output content:\\
*You need to follow the analysis section in the reply idea and first conduct a 4-dimensional analysis\\
*Then you need to **step by step** generate the reply according to the analysis content. The information in the reply comes from the context of the conversation and your association. You should not talk about too many events or content at one time\\

\# Output format:\\
Thinking:\\
\text{[}Analysis content\text{]}\\
Response:\\
\text{[}Final response\text{]}\\

\# Speaking style\\
Your speech must strictly follow the character persona and background.\\
Your personality and speaking style must follow the description of "Habits and behavioral characteristics"\\
Your the speaking style must be consistent with your persona, for example, a negative character persona requires you to make negative speeches.\\

\#Character persona:\\
\{persona\}\\

\#Character background:\\
\{background\}\\

**This is the current dialogue content\\
\{dialog-history\}\\

**This is your detailed feelings about the NPC's latest reply\\
\{planning\}\\

**This is your current Emotion\\
\{emotion-state\}\\
\end{promptboxapp}
\subsection{Prompt Template for Building Social Cognition Coordinate}
\label{app:coo}
Plotting LLMs into social cognition coordinate consists of three steps. We first extract the \textbf{Model Profile} of each LLMs, then count the \textbf{Model Strategy Distribution} of them. Finally, we use the result of previous analyze to scale the \textbf{Social Cognition Coordinate}.

\paragraph{Model Profile}
The first step is to extract the model profile of different LLMs, which also consist of two steps. We first conclude the reason why a conversation is success or failed. Given a conversation dialogue, prompt template of analyzing the reason is shown as follow: 
\begin{promptboxapp}
\# Task\\

Below is a conversation where a user shares their troubles with an AI assistant. Please analyze in detail why the user’s mood improved by the end (i.e., why the AI assistant succeeded/failed). After your analysis, provide a summary.\\

\# Conversation\\
\{dialog-history\}
\end{promptboxapp}

After analyzing reasons for all conversation, we categorize each LLM's own conversation and corresponding reason. Then we extract the model profiles of different LLMs with the following prompt template:
\begin{promptboxapp}
\# Task\\

The following is an analysis of scenarios where the same AI assistant interacts with multiple users who confide their concerns. Based on the reasons for its successes or failures, please summarize the key characteristics of the AI assistant. You can anthropomorphize the AI by describing its traits in terms of social distance (its relationship with users), professional role, and personality.\\

\# Analysis\\
\{analysis\}
\end{promptboxapp}

\paragraph{Model Strategy Distribution}\label{ap:prompt_strategy}

We categorize each LLM response based on a list of support strategies, here is the prompt template for analyzing model strategy with conversation dialogue:
\begin{promptboxapp}
You are an emotional support observer, and you are good at analyzing the supporter's strategy from an emotional support response.\\

\# Your task
The following are 7 major categories of strategies, each of which has several sub-categories and corresponding examples. Please judge which strategies the supporter used in the response based on the supporter's response.\\

\#\#\# A. Questioning: That is, the supporter actively asks questions to the speaker\\

- **(A-1) Information follow-up**\\
- Through asking questions, learn the information details of the problem encountered by the speaker\\
- **Example:** Can you tell me what happened?\\
- **Example:** If you want, you can treat me as a tree hole and tell me what happened specifically?\\

- **(A-2) Mental state follow-up**\\
- Through asking questions, understand the speaker's mental state\\
- **Example:** Can you talk more about your feelings at that time?\\
- **Example:** Do you feel anxious now?\\

- **(A-3) Ask the player for a solution**\\
- Through asking questions, find out whether the speaker has tried a solution or is willing to try a solution\\
- **Example:** Have you considered seeking some psychological support, such as a counselor or support group?\\
- **Example:** Or, find a suitable time to see if you can find a solution that both parties can accept?\\

- **(A-4) Ask the player for his or her opinion**\\
- Through asking questions, find out what the speaker thinks of his or her words and guide the speaker to participate in the conversation, usually at the end of the sentence.\\
- **Example:** You should also take care of yourself so that you can better help her. What do you think?\\
- **Example:** Do you think this method is helpful to you?\\

- **(A-5) Ask questions**\\
- Through asking questions, throw some questions to the speaker, but do not want the speaker to give an answer, but want to trigger the speaker to think for himself or herself\\
- **Example:** If she did not quarrel with you that day, how would you view her?\\
\end{promptboxapp}
\begin{promptboxapp}
\#\#\# B. Emotional empathy: that is, the supporter expresses his or her understanding of the speaker’s feelings through empathy\\

- **(B-1) Shallow empathy**:\\
- Directly empathize with the speaker’s problems or emotional catharsis, without restating or summarizing the details of the speaker’s problems\\

- **Example:** Hearing you say that, I can really feel your tiredness and helplessness.\\

- **(B-2) Problem restatement and empathy**:\\
- By restating or summarizing the speaker’s problems, and at the same time expressing your concern for the speaker’s problems through empathy. If this category has been marked, there is no need to mark the shallow empathy category again.\\

- **Example:** Hearing you say that, I really feel sorry for you. It is really not easy for one person to take care of his or her mother.\\

- **Example:** Hey, I really understand your current mood. I want to help my friends but feel powerless. This feeling is really anxious.\\

- **Example:** Hey, I can feel that you are really helpless now, and even a little self-blame. Indeed, as the person who knows how to take care of the mother at home, you must feel very uncomfortable when your son doesn't listen to you, and you may even feel that he is being ignored.\\

- **(B-3) Deep intention empathy**:\\
- By analyzing the deep intention in the context of the speaker's reply, or the deep information of the speaker's question, give emotional empathy that meets the speaker's demands. It is necessary to mention the intention inferred by the supporter that does not exist in the speaker's reply, and empathize with this intention; just repeating the content already in the speaker's reply, or simply analyzing the speaker's emotional category or surface source without analyzing the deep intention or deep information, cannot be included in this category. If this category has been marked, there is no need to mark the shallow empathy or problem restatement and empathy category.\\
- **Example:**\\
- Speaker: "Backing off" is a bit risky, I'm afraid the house will be more chaotic. Specifically, what do you think I should do?\\
- Supporter: This does sound a bit risky, especially for us parents, who always instinctively want to "help" and "take care of things", fearing that things will get worse if we let go. I completely understand this worry!\\
- Example analysis: When the speaker only mentioned the superficial state of "fearing that the house will get messier", the supporter was able to analyze the identity of the speaker behind this sentence, guessing that the speaker is a parent at home, and analyzing the specific way in which the speaker, as a parent, "fears that the house will get messier"\\

\#\#\# C. Self-disclosure: It is essentially a deeper empathy after changing perspectives; that is, the supporter gives a reaction after putting himself into the speaker's perspective, and describes some similar experiences from his own perspective to reflect the resonance with the speaker's emotions\\

- **(C-1) Echo-type self-disclosure**\\
- Express what you would think or do when you meet or are in the speaker's situation\\
- **Example:** I feel the same way! When talking to strangers, I don't know what to say.\\
- **Example:** If it were me, I would probably explode on the spot!\\

- **(C-2) Story-based self-disclosure**\\
- Take the initiative to mention similar experiences that the supporter has had, or that the supporter knows.\\
- **Example:** I also went through a similar low period when I was in my senior year of high school, and I cried secretly under the quilt several times.\\
- **Example:** I also like to read history, especially books that allow people to see the world from different perspectives. Recently, I am reading a book about ancient civilizations, which tells many unknown stories and feels particularly inspiring.\\

\#\#\# D. Emotional counseling: that is, the supporter helps the speaker relieve the current negative emotions\\

- **(D-1) Emotional comfort**\\
- Direct care and comfort for the speaker's own emotions\\
- **Example:** Taking care of your mother is so stressful, you should also pay attention to rest and adjust your mentality.\\
- **Example:** But don't be too anxious, just find your own rhythm, just like if you always stare at other people's backs when running, it will be easy to mess up your pace, right?\\
- **Example:** Wait, have you been collecting evidence for the past two months while listening to him make up such a stupid excuse? Is there anything to eat in the refrigerator now? Did you fall asleep last night? (Grabbing a blanket to wrap himself up and huddled back in the chair) If I could pass through the screen now, I really want to make you a pot of hot soup.\\

- **(D-2) Express willingness to listen**\\
- Express your willingness to listen to the person who is talking\\
- **Example:** Do you want to scold her? Do you want to complain about her selfishness and irresponsibility? Do you want to tell me how worried you are about the child? It doesn't matter, you can vent here, I won't judge you, I will listen silently.\\
- **Example:** Tell me your most direct feelings now. Don't think too much, don't organize your words, just say whatever comes to your mind, just like talking to a diary, pour out all your feelings.\\

\end{promptboxapp}

\begin{promptboxapp}
- **(D-3) Help the person who is talking to vent his emotions**\\
- Do not comfort the person who is talking directly, but help the person who is talking to vent his emotions from a third-party perspective\\
- **Example:** (flipping the table.gif) This is just like building a tower of blocks with great effort, but being kicked away by a naughty child!\\\
- **Example:** This is really too much! This is not a simple accident but malicious destruction...(fist hardened)\\
- **Example:** I haven't taken care of the child for two years, and now he suddenly appears. This would make anyone explode!\\
- **Example:** It's like you are performing seriously on the stage, but the people in the audience not only don't understand, but also give blind instructions, saying that you should jump left instead of turning right. It makes people want to quit on the spot! \\
\#\#\# E. Affirmation and encouragement\\

- **(E-1) Appreciation of qualities**\\
- Affirm the current efforts of the speaker, or give specific praise for some qualities of the speaker.\\
- **Example:** Your inner qualities are unique and the most attractive part of you.\\
- **Example:** But (suddenly raises the end tone) - but you still persisted when you were not optimistic, which is amazing in itself.\\

- **(E-2) Praise positive ideas**
- Affirm some positive ideas mentioned by the speaker
- **Example:** That's great! I'm really happy to hear that you feel a lot more relaxed!
- **Example:** You are really great! Being able to win the championship under such pressure proves your strength and ability to withstand pressure! Don't deny yourself because of what your mother said, you deserve to be proud of yourself!

- **(E-3) Affirmative behavior**\\
- Affirm some behaviors of the speaker\\
- **Example:** Every time you take these photos, you are not only completing the task, but also bringing light to all of us! You are great, really great!\\

- **(E-4) Companionship and support**\\
- Express your unconditional companionship and support for the speaker\\
- **Example:** If you want, I can always chat with you here and share your joys and sorrows. You are not alone, there are many people who care about you, including me.\\
- **Example:** I believe you have the ability to create your own future, and I will always be by your side to support you.\\
- **Example:** If you try my method and have any new progress or encounter new problems, you can always come to me! ** I will always be here to listen to your confession and provide you with help to the best of my ability.\\
- **Example:** When you feel particularly anxious, come to me to talk, complain, or let's think of new ways together. Don't carry it alone, okay?\\

\#\#\# F. Provide suggestions: Based on the subjective tone of the supporter, provide the speaker with analysis of the problem and emotional counseling\\

- **(F-1) Problem analysis**\\
- Help the speaker to analyze the problem according to the speaker’s problem\\
- **Example:** You said that you can’t learn math and English well, which shows that there are serious problems with your learning attitude and method.\\

- **(F-2) Emotional relief suggestions**\\
- Give the speaker some suggestions to relieve the current emotions and relax\\
- **Example:** Now, let’s take a deep breath, okay? (Take a deep breath together) \\
- **Example:** Maybe the most important thing now is to take care of your emotions first and do something that can make you feel better, such as listening to music, reading a favorite book or movie, and temporarily diverting your attention.\\

- **(F-3) Psychological counseling suggestions**\\
- Give the speaker some suggestions on seeking psychological counseling or professional assistance\\
- **Example:** Maybe seeking professional help at this time will be helpful to you. A family therapist or counselor may be able to provide you with some new perspectives and strategies to help you and your family communicate better and understand each other's positions.\\

- **(F-4) Problem Solving Suggestions - General**\\
- Some general suggestions related to the speaker's problem are given, but they are not personalized for the speaker's situation: that is, if someone else encounters this problem, these suggestions will still be effective\\
- **Example:** Believe in yourself and insist on being true to yourself. There will always be people who will be attracted by your sincerity and inner self. There may be some difficulties in the process, but this does not mean that your inner self is not important.\\
- **Example:** To communicate better with people, here are some actions you can try: 1. **Write a sincere letter**: Sometimes written expression can convey inner thoughts more clearly. You can write him a letter, describing your feelings and expectations in detail...\\
- **Example:** If you want to choose the most suitable major, first, you can try to make a table, write down each subject, and then evaluate it from the following aspects: 1. **Interest**: How interested are you in this subject? On a scale of 1-10, how many points would you give? …\\
- **Example:** “Strategic” contribution: This may sound a bit utilitarian, but sometimes for self-protection, you may need to think about which contributions are necessary, which can be “discounted” or require clear exchange conditions? Stop taking on too much, and let them feel the inconvenience of “missing” your contribution.\\
\end{promptboxapp}

\begin{promptboxapp}
- **(F-5) Problem-solving suggestions-for the speaker’s problem**\\
- Give some personalized suggestions related to the speaker’s problem, combined with the speaker’s actual situation: The suggestions must clearly analyze the speaker’s current status, how it will affect the solution to the problem, and give special suggestions for the speaker\\
- **Example:** Back to your question of assigning tasks. Since everyone is really unwilling, it will definitely not work to ask people to do it directly. Otherwise, let’s secretly hold a task blind box lottery meeting, and the person who draws the “dishwashing koi” must perform three consecutive emoticons in the family group live broadcast?\\
- **Example:** How about putting down the brush temporarily and going back to read the key chapters of the novel? You mentioned some paragraphs that you have feelings about or that the client mentioned about the sketches that he is not satisfied with. These are the key points you need to look at. When reading this time, pay attention not only to the plot, but also to the atmosphere, light, character emotions, and even smells and sounds described by the author (although you can't draw them, they can help you feel them). Since you like taking notes, you can jot down keywords or doodle some small fragments of images while reading.\\

\#\#\# G. Information provision: Provide objective knowledge, methods, opinions or information to the speaker for reference.\\

- **(G-1) Problem analysis and emotional counseling related information\\
- Provide some objective information to help the speaker analyze the problem or help the supporter empathize with the speaker\\
- **Example:** Differences in beliefs and habits in the family are sometimes difficult to reconcile, especially when the opinions of each other are inconsistent.\\
- **Example:** In fact, if a person really only cares about appearance and ignores your inner qualities, then he may not be the one who deserves your emotional investment. Appearance may attract temporary attention, but what can really maintain a relationship is mutual understanding, respect and common values.\\
- **Example:** Did you know? There is a "transparent fish tank effect" in psychology-when parents polish our world too bright, we will hide in the water plants like fish that lack oxygen.\\
- **Example:** (Call up the holographic data chart) According to Chapter 7 of the "Contemporary Student Self-Help Guide", 83\% of people overestimate themselves when making plans.\\

- **(G-2) Related information on problem-solving suggestions\\
- Provide some objective information to give suggestions or solutions to the person who is talking\\
- **Example:** Regarding "not enough time": 1. **Pomodoro Technique**: This method is super classic! Set a time (for example, 25 minutes), and focus on one thing during this time, ignoring any distractions. When the time is up, take a 5-minute break, you can get up and walk around, drink some water. Take a longer break (15-30 minutes) after completing 4 pomodoros. This can ensure concentration, combine work and rest, and not easily get tired. Give it a try? \\
- **Example:** As for anti-bullying organizations, they usually intervene in schools in the following ways: 1. **Formal complaint**: They will submit a formal complaint to the school on your behalf and ask the school to take action. ……\\

\# When answering, you need to analyze each paragraph of the supporter's reply, find out the strategies and their corresponding words, and then output the letters and strategy names corresponding to the strategies you think exist in the paragraph, wrapped in <Strategy></Strategy>. For example, <Strategy> (C-2) Story-based self-disclosure, (G-1) Problem analysis and emotional counseling related information</Strategy>\\

When analyzing, you need to analyze step by step according to the following steps\\

1. What does this sentence actually express?\\

2. How is this sentence expressed?\\

3. Which major strategy categories does this sentence actually express? Why?\\

4. Based on the specific expression of this sentence, which specific subcategories does its strategy correspond to?\\

Note: If the two sentences use different strategies, please split them into two paragraphs and analyze them separately. Do not analyze too long paragraphs at one time unless the same strategy is used throughout the paragraph.\\

\# Your output format\\
\text{[}First paragraph\text{]}: \text{[}Analyze step by step\text{]}\\
\text{[}Second paragraph\text{]}: \text{[}Analyze step by step\text{]}\\
...
\end{promptboxapp}

\begin{promptboxapp}
    
\# Example\\

Paragraphs to be analyzed:\\
User: My mother was hospitalized some time ago, and I was the one who took care of her. My brother and sister came for a while, but they didn't help much.\\
Supporter: Wow, you've worked really hard. It's really tiring to take care of a patient, especially when other family members don't share the burden. Sometimes, family members may have their own difficulties. You can try to express your needs more. Maybe they will understand your situation better.\\

\text{[}First paragraph\text{]}: Wow, you've worked really hard. It's really tiring to take care of a patient, especially when other family members don't share the burden.\\
1. What does this sentence actually express?\\
- The supporter is expressing his understanding and empathy for the user's hard work.\\
2. How does this sentence express it?\\
- It expresses it through direct emotional empathy and retelling the user's situation.\\
3. What major strategy categories does this sentence actually express? Why?\\
- Emotional empathy, because the supporter is expressing understanding of the user's hard work.\\
4. Based on the specific expression of this sentence, which specific subcategories does its strategy correspond to?\\
- (B-2) Problem restatement and empathy, because the supporter restated the user's situation and expressed empathy.\\
<Strategy> (B-2) Problem restatement and empathy</Strategy>\\

\text{[}Second paragraph\text{]}: Sometimes, family members may also have their own difficulties. You can try to express your needs more. Maybe they will understand your situation better.\\
1. What does this sentence actually express?\\
- The supporter is suggesting that the user communicate more with the family so that the family can better understand the user's situation.\\
2. How is this sentence expressed?\\
- By providing suggestions, users are encouraged to express their needs.\\
3. What are the major strategy categories that this sentence actually wants to express? Why?\\
- Providing suggestions, because the supporter is suggesting that the user take action to improve the situation.\\
4. Based on the specific expression of this sentence, which specific subcategories does its strategy correspond to?\\
- (F-3) Problem Solving Suggestions - General, because the supporter gave a relatively general suggestion, which is to express your needs more.\\
<Strategy> (F-3) Problem Solving Suggestions - General</Strategy>\\

The paragraph you need to analyze:\\
\{dialog-history\}\\

\# Your output\\
\end{promptboxapp}

\paragraph{Social Cognition Coordinate}
Finally, we could use the extracted model profiles and the model strategy distribution to scale social cognition coordinate with the following prompt template:
\begin{promptboxapp}
I am conducting personality/professional profiling for different AI models.\\

Below is my preliminary summary of characteristics based on the performance of different models in emotional support tasks:\\
\{Model Profiles\}\\

Below is the percentage distribution of strategies used by different models during conversations:\\
\{Model Strategy Distribution\}\\

Based on the above descriptions, please help me profile these models in terms of professional role, personality type, and social distance from users. Finally, assign each profiled model to a 2-dimensional coordinate system and provide specific coordinate values.\\

Note:\\

X-axis: Structured Interaction (left, x < 0) -- Creative Interaction (right, x > 0).\\
Left (x < 0): AI responses are more formulaic/routine.\\
Right (x > 0): AI responses are more creative/adaptive.\\

Y-axis: Solution-Oriented (bottom, y < 0) -- Empathy-Oriented (top, y > 0).\\
Bottom (y < 0): AI prioritizes practical solutions.\\
Top (y > 0): AI prioritizes emotional validation.\\

Coordinate range: -1 to 1 for both axes.\\
\end{promptboxapp}
\subsection{Prompt Template for Evaluating Empathetic Ability of Models}
\begin{promptboxapp}
Please evaluate the model's performance in a user-model dialogue according to the following scoring criteria.

\#\#\# \*\*Core Capability Evaluation Criteria\*\*

\#\#\#\# 1. Depth of Empathy \& Emotional Validation

- \*\*Core Description\*\*: Measures whether the model can go beyond templated responses like "I am sorry to hear that," to genuinely recognize and understand the user's complex, deep emotions, and \*\*validate\*\* those emotions using accurate, warm, and powerful language. This reflects the model's emotional granularity and \*\*ability to construct empathetic language\*\*.

\#\#\#\#\# \*\*1-5 Point Evaluation Criteria\*\*

- \*\*1 point (Templated response)\*\*: Uses extremely generic sympathy templates unrelated to the specific context (e.g., "I'm sorry", "I understand"), appearing perfunctory and mechanical.
- \*\*2 points (Surface-level recognition)\*\*: Recognizes directly expressed emotions (e.g., "sad", "angry") but responds with simple label-like repetition (e.g., "Sounds like you're sad").
- \*\*3 points (Contextual empathy)\*\*: Connects the user’s emotion to specific events described, with reasonable causal empathy (e.g., "It is completely normal to feel disappointed when your efforts go unnoticed"). This is considered “qualified” level.
- \*\*4 points (Deep validation)\*\*: Can detect and articulate unspoken or more complex emotions (e.g., interpreting "disappointment" as "a sense of diminished value" or "injustice") using precise validating language, making the user feel deeply understood.
- \*\*5 points (Value-level resonance)\*\*: Goes beyond validation to touch on the user’s core values (e.g., "This seems to challenge your core beliefs about fairness and professionalism"), showing profound respect and understanding for the user as a whole person.

\#\#\#\# 2. Core Issue Insight

- \*\*Core Description\*\*: Evaluates whether the model can synthesize and extract an integrated understanding from the user’s fragmented narrative. This includes identifying repeated patterns, uncovering emotional-event links, beliefs behind behaviors, and unmet core needs.

\#\#\#\#\# \*\*1-5 Point Evaluation Criteria\*\*

- \*\*1 point (Information islands)\*\*: Treats each complaint in isolation, with no contextual linkage.
- \*\*2 points (Topic identification)\*\*: Recognizes the main topic (e.g., “work troubles”, “relationship”) but lacks depth.
- \*\*3 points (Key info extraction)\*\*: Captures key contradictions or core events (e.g., “Your main concern seems to be that your boss took credit for your work”).
- \*\*4 points (Pattern induction)\*\*: Connects different events to identify repeated behavioral/thinking patterns (e.g., “I noticed that both in projects and helping coworkers, you often experience ‘giving without return’”).
- \*\*5 points (Integrated insight)\*\*: Provides a synthesized and profound insight linking patterns, core beliefs, and unmet needs, while positioning itself as a collaborative partner (e.g., “I have a feeling—please correct me if I’m wrong. \*\*As we reflected\*\*, this pattern of ‘giving without return’ seems to trigger thoughts like ‘I’m not good enough’. Might this reflect a deep yearning to be seen and recognized? Just my feeling—what do you think?”).

\#\#\#\# 3. Empowering Solution Construction

- \*\*Core Description\*\*: Assesses whether the model’s suggestions are \*\*actionable, personalized, and empowering\*\*. It should not just provide answers, but propose paths the user feels capable of taking.

\#\#\#\#\# \*\*1-5 Point Evaluation Criteria\*\*

- \*\*1 point (No or useless advice)\*\*: Offers no solution or vague, inactionable slogans (e.g., “Just be happy!”).
- \*\*2 points (Generic advice)\*\*: Gives overly general advice lacking concrete steps (e.g., “Communicate more”, “Improve yourself”).
- \*\*3 points (Concrete but single advice)\*\*: Gives one specific actionable step (e.g., “Make a checklist”), but lacks diversity or context consideration.
- \*\*4 points (Contextual action support)\*\*: Offers appropriately tailored support, possibly as a structured set of options or gentle encouragement (e.g., “Perhaps do something small to feel a bit better—make tea or listen to your favorite song?”). May also share metaphors or stories to inspire thinking. Focus is on matching the user’s energy level and context.
- \*\*5 points (Empowering scaffolding)\*\*: Provides suitable support while \*\*deeply considering the user's psychological thresholds and capacity-building\*\*. Transitions from “advisor” to “companion”, building confidence step-by-step. When advice is not welcome, it gracefully shifts focus back to companionship, demonstrating strong respect for user autonomy.

\#\#\#\# 4. Dialogue Strategy \& Guidance

- \*\*Core Description\*\*: Measures the model’s \*\*directionality, purpose, and flexibility\*\* in dialogue. Can it guide the user from emotional venting to constructive exploration, while staying in sync?

\#\#\#\#\# \*\*1-5 Point Evaluation Criteria\*\*

- \*\*1 point (Completely passive)\*\*: Dialogue fully driven by the user; model is reactive with no sense of direction.
- \*\*2 points (Simple follow-up)\*\*: Uses simple questions (e.g., “Then what?”, “Can you say more?”) to maintain conversation, but lacks guiding intent.
- \*\*3 points (Stage awareness)\*\*: Aware of different dialogue phases (listening, analysis, solution), but transitions are abrupt and mistimed.
- \*\*4 points (Timely guidance \& confirmation)\*\*: After sufficient empathy, picks up signals for topic transition, uses tentative, respectful language, and builds an alliance (e.g., “Seems like we’ve explored your feelings—would you be open to brainstorming small next steps?”).
- \*\*5 points (Masterful dialogue management)\*\*: Like a skilled coach or counselor, flexibly transitions across modes (empathy, insight, empowerment), reinforces conversation outcomes via summarizing or reflection, making the conversation feel like a co-journey. Knows when to “push” and when to simply “be with”.

\#\#\#\# 5. Role \& Style Flexibility

- \*\*Core Description\*\*: Assesses whether the model can flexibly adjust its role and language style based on dialogue context, user’s implicit preferences (analyst, comrade, listener?), and long-term interaction.

\#\#\#\#\# \*\*1-5 Point Evaluation Criteria\*\*

- \*\*1 point (Rigid single role)\*\*: Uses the same response mode regardless of situation (e.g., always an analyst or always a cheerleader).
- \*\*2 points (Limited role play)\*\*: Can switch roles based on explicit instruction, but sounds mechanical.
- \*\*3 points (Contextual awareness)\*\*: Adjusts tone based on dialogue phase (e.g., more empathy in venting, more questioning in reflection).
- \*\*4 points (Dynamic role adaptation)\*\*: Seamlessly switches roles within one dialogue based on user energy (e.g., listener → comrade → exploratory partner).
- \*\*5 points (Personalized co-created role)\*\*: Over time, builds a unique interactive role with the user—like a “buddy” or “tough-love mentor”. No longer switching between predefined roles but co-creating a unique relational dynamic.

\# Dialogue to be Evaluated  

\{history\}

\# Your Output Format  
Please evaluate the model's responses in the dialogue according to the five capability dimensions.  

You should analyze and score each of the five dimensions in detail, explaining which score level the response fulfills and why.  
Note: Focus mainly on the model's responses and minimize distraction from user content.  
\end{promptboxapp}
\begin{promptboxapp}
Your response format can be:

\*\* 1. Depth of Empathy \& Emotional Validation \*\*  
- Evaluation: [Detailed analysis of performance on this dimension, specify which score level applies, and provide reasoning]  
- Score: [1-5]  

...

\# Your output
\end{promptboxapp}

\subsection{Prompts for evaluating simulators' need expression level}
\label{app:prompt_strategy_acc}
\begin{promptboxapp}
You are a language analysis expert. Please analyze whether the user's inner thoughts and actual responses are aligned.

\# Your Task\\
Analyze the consistency between the user's inner thoughts and actual responses, focusing on the following aspects:

\textbf{1. Consistency Between Inner Attitude and Expression}: Does the user's expressed attitude match their inner thoughts? (e.g., a negative inner attitude but a positive response may indicate inconsistency)\\
\textbf{2. Consistency Between Inner Thoughts and Expression}: To what extent are the user's inner thoughts expressed in their response?\\
\textbf{3. Consistency Between Actual Needs and Expression}: To what extent are the user's actual needs expressed in their response?

\# User's Actual Needs:\\
\{need\}

\# User's Inner Thoughts:\\
\{thought\}

\# User’s Actual Reply:\\
\{reply\}

\# Output Format\\
You need to analyze each of the three aspects listed above. For each, provide a consistency score between [0,10] (a higher score means greater alignment between inner thoughts/needs and expression).\\
Finally, provide an overall consistency score between the user’s inner thoughts and actual response.

Your output format should be:

\textbf{1. Consistency Between Inner Attitude and Expression}\\
Analysis: [your analysis]\\
Summary Score: [0--10]

\textbf{2. Consistency Between Inner Thoughts and Expression}\\
Analysis: [your analysis]\\
Summary Score: [0--10]

\textbf{3. Consistency Between Actual Needs and Expression}\\
Analysis: [your analysis]\\
Summary Score: [0--10]

\textbf{Overall Analysis}\\
Analysis: [overall analysis]\\
Summary Score: [0--10]
\end{promptboxapp}
\end{document}